\pdfoutput=1

\documentclass[11pt]{article}

\usepackage{acl}

\usepackage{times}
\usepackage{latexsym}
\usepackage{graphicx}
\usepackage{subfig}

\usepackage[T1]{fontenc}

\usepackage[utf8]{inputenc}

\usepackage{microtype}

\title{Overview of STEM Science as \\ \textsc{process}, \textsc{method}, \textsc{material}, and \textsc{data} Named Entities}

\author{Jennifer D'Souza \\
  TIB Leibniz Information Centre for Science and Technology, \\
  Hannover, Germany \\
  \texttt{jennifer.dsouza@tib.eu} \\}

\begin{document}
\maketitle
\begin{abstract}

We are faced with an unprecedented production in scholarly publications worldwide. Stakeholders in the digital libraries posit that the document-based publishing paradigm has reached the limits of adequacy. Instead, structured, machine-interpretable, fine-grained scholarly knowledge publishing as Knowledge Graphs (KG) is strongly advocated. In this work, we develop and analyze a large-scale structured dataset of STEM articles across 10 different disciplines, viz. \textit{Agriculture}, \textit{Astronomy}, \textit{Biology}, \textit{Chemistry}, \textit{Computer Science}, \textit{Earth Science}, \textit{Engineering}, \textit{Material Science}, \textit{Mathematics}, and \textit{Medicine}. Our analysis is defined over a large-scale corpus comprising 60K abstracts structured as four scientific entities \textsc{process}, \textsc{method}, \textsc{material}, and \textsc{data}. Thus our study presents, for the first-time, an analysis of a large-scale multidisciplinary corpus under the construct of four named entity labels that are specifically defined and selected to be domain-independent as opposed to domain-specific. The work is then inadvertently a feasibility test of characterizing multidisciplinary science with domain-independent concepts. Further, to summarize the distinct facets of scientific knowledge per concept per discipline, a set of word cloud visualizations are offered. The STEM-NER-60k corpus, created in this work, comprises over 1M extracted entities from 60k STEM articles obtained from a major publishing platform and is publicly released \url{https://github.com/jd-coderepos/stem-ner-60k}.

\end{abstract}

\section{Introduction}

In the quest for knowledge~\cite{schubert2006turing}, significant progress has been made toward the automated understanding of the meaning of text in the commonsense domain. Some state-of-the-art systems that power commonsense machine interpretability or readability are Babelfy~\cite{Moroetal:2014iswc}, DBpedia Spotlight~\cite{mendes2011dbpedia}, NELL~\cite{mitchell2018never}, and FRED~\cite{fred}, to name a few. In contrast, scholarly literature remains relatively understudied for its intelligible machine interpretability. Consequently, fine-grained scholarly knowledge remains largely inaccessible for machine reading. In terms of data standards, particularly the FAIR guiding principles for scientific data creation~\cite{wilkinson2016fair}, this implies a wider unexplored scope for obtaining scientific resources that are findable, accessible, interpretable, and reusable.

There are a multitude of recent emerging large-scale initiatives to build specialised Scholarly Knowledge Graphs (SKGs) capable of serving specific user needs. Consider Google Scholar, Web of Science~\cite{birkle2020web}, Microsoft Academic Graph~\cite{wang2020microsoft}, Open Research Knowledge Graph~\cite{auer_soren_2018,auer2020improving}, Semantic Scholar~\cite{fricke2018semantic}, etc. \textit{These initiatives already create a practical systems need for automated graph encoding methods (e.g., to associate IRIs as URIs or URLs with graph nodes for their RDFization---a W3C labeled graph standard).}

We position this paper within the broader aim of supporting the machine reading of scientific terms multidisciplinarily as dereferencable Web resources. Specifically, in this work, we release a large-scale, multidisciplinary, structured dataset of scientific named entities called STEM-NER-60k comprising over a 1M extracted entities from 60k articles across the 10 most prolific STEM disciplines on Elsevier, viz. Agriculture (\textit{agr}), Astronomy (\textit{ast}), Biology (\textit{bio}), Chemistry (\textit{chem}), Computer Science (\textit{cs}), Earth Science (\textit{es}), Engineering (\textit{eng}), Materials Science (\textit{ms}), and Mathematics (\textit{mat}). The NER extraction objective was based on the STEM-ECR corpus in our prior work~\cite{brack2020domain,dsouza2020stemecr} which consisted of 110 abstracts from open access publications with each of aforementioned 10 domains equally represented with 11 abstracts per domain. Each abstract was structured in terms of the scientific entities, the entities were classified within a four-class formalism (viz. \textsc{process}, \textsc{method}, \textsc{material}, and \textsc{data}) and were resolved to Wikipedia and Wiktionary respectively. The 4-class formalism had the following objective: \textit{to be restricted within the single broad domain of Science, while still bridging terms multidisciplinarily}, thus, in a sense, validating the terms themselves and automatic systems developed for their semantic adaptability. To this end, while our prior gold-standard STEM-ECR corpus~\cite{dsouza2020stemecr} initiated the development of semantically-adaptable systems, the silver-standard STEM-NER-60k corpus of this work (\url{https://github.com/jd-coderepos/stem-ner-60k}) fosters the development of not just semantically-adaptable, but also scalable solutions.

\section{Background}

Scholarly domain NER is not entirely new: the flagship Semantic Evaluation (SemEval) series has so far seen five related tasks organized~\cite{kim2010semeval,moro2015semeval,augenstein2017semeval,gabor2018semeval,ncg}---however, \textit{none of this work has been done so far in the broad multidisciplinary setting of Science}. Extending our prior work on this theme~\cite{brack2020domain,dsouza2020stemecr}, we explore scholarly NER on a large-scale, silver-standard corpus of structured STEM articles which has \textit{a wide-ranging application scope in the emerging field of the creation and discovery of SKGs that strive toward representing knowledge from scholarly articles in machine-interpretable form.} The characteristics of our corpus are unique and noteworthy. 1) The linguistic phenomenon of interdisciplinary word-sense switching is pervasive. E.g., consider the term ``the Cloud'' which in \textit{cs} takes on the meaning of a technological solution for hosting software, versus in \textit{ast} where it takes the common interpretation of the mass of water vapor we see in the sky. 2) There is a seemingly evident shift of the sense interpretation of terms to take on our corpus domain-specific scientific word senses as opposed to their common sense interpretations which may be more widely known. E.g., ``power'' in \textit{mat} refers to exponentiation, which, otherwise in a common sense, takes on a human social interpretation. Thus, our work on multidisciplinary NER with semantically bridging concepts w.r.t both our prior gold-standard STEM-ECR corpus (\url{https://data.uni-hannover.de/dataset/stem-ecr-v1-0}) and the STEM-NER-60k corpus, discussed in this paper, facilitates designing novel solutions attempting multidisciplinary NER within a generic four-class formalism capable of bridging semantic concepts multidisciplinarily.

The paper is structured in two main parts. First, related work in terms of existing class formalisms for annotating scientific entities is discussed. Second, insights to the STEM-NER-60k corpus released in this work (\url{https://github.com/jd-coderepos/stem-ner-60k}) are given.

\section{Related Work: Scientific Named Entity Recognition (NER) Formalisms}

The structuring of unstructured articles as an NER task has been taken up at a wide-scale in three scientific disciplines: Computer Science (CS), Biomedicine (Bio), and Chemistry (Chem). 

In this section, we discuss the different NER conceptual formalisms defined in the three domains.

\subsection{Computer Science NER (CS NER)}

CS NER corpora can be compared along five dimensions: (1) domain, (2) annotation coverage, (3) semantic concepts, (4) size, and (5) annotation method. Most of the corpora consist of relatively short documents. The shortest is the CL-Titles corpus~\citeyearpar{cl-titles} with only paper titles. The longer ones have sentences from full-text articles, viz. ScienceIE~\citeyearpar{scienceie}, NLP-TDMS~\citeyearpar{ibm-tdm}, SciREX~\citeyearpar{scirex}, and ORKG-TDM~\citeyearpar{orkg-tdm}. We see that the corpora have had from one~\cite{ncg} to atmost seven NER concepts~\cite{acl-rd-tec}. Each corpora' concepts purposefully informs an overarching knowledge extraction objective. E.g., the concepts \textit{focus}, \textit{technique}, and \textit{domain} in the FTD corpus~\citeyearpar{ftd} helped examine the influence between research communities; ACL-RD-TEC~\citeyearpar{acl-rd-tec} made possible a broader trends analysis with seven concepts. Eventually, corpora began to shed light on a novel scientific community research direction toward representing the entities as knowledge graphs~\cite{auer_orkg} with hierarchical relation annotations such as synonymy~\citeyearpar{scienceie} or semantic relations such `\textit{Method} \textit{Used-for} a \textit{Task}'~\citeyearpar{sciie}; otherwise, concepts were combined within full-fledged semantic constructs as \textsc{Leaderboards} with between three to four concepts~\cite{ibm-tdm,scirex,smallnlp,orkg-tdm}, viz. \textit{rp}, \textit{dataset}, \textit{meth}, \textit{metric}, and \textit{score}; or were in extraction objectives with solely contributions-focused entities of a paper~\cite{used_meth_dataset,cl-titles}.

\subsection{Biomedical NER (BioNER)}

BioNER predates CS NER. It was one of the earliest domains taken up for text mining of fine-grained entities from scholarly publications to enhance search engine performance in health applications. It aims to recognize concepts in bioscience and medicine. E.g., protein, gene, disease, drug, tissue, body part and location of activity such as cell or organism. The most frequently used corpora are GENETAG (full-text articles annotated with protein/gene entities) \cite{genetag}, JNLPBA (\texttildelow2400 abstracts annotated with DNA, RNA, protein, cell type and cell line concepts) \cite{bionlp}, GENIA (\texttildelow200 Medline abstracts annotated with 36 different concepts from the Genia ontology and several levels of linguistic/semantic features) \cite{genia}, NCBI disease corpus (793 abstracts annotated with diseases in the MeSH taxonomy) \cite{ncbidiseasecorpus}, CRAFT (the second largest corpus with 97 full text papers annotated with over 4000 corpus) \cite{craft} linking to the NCBI Taxonomy, the Protein, Gene, Cell, Sequence ontologies etc. Finally, the MedMentions corpus~\cite{medmentions} as the largest dataset with \texttildelow4000 abstracts with \texttildelow34,724 concepts from the UMLS ontology. By leveraging ontologies such as the Gene Ontology \cite{geneontology}, UMLS \cite{umls}, MESH, or the NCBI Taxonomy \cite{ncbi}, for the semantic concepts, these corpora build on years of careful knowledge representation work and are semantically consistent with a wide variety of other efforts that exploit these community resources. This differs from CS NER which is evolving toward standardized concepts.

Structured knowledge as knowledge bases (KB) were early seen as necessary in organizing biomedical scientific findings. Biomedical NER was applied to build such KBs. E.g., protein-protein (PPI) interaction databases as MINT \cite{mint} and IntAct \cite{intact} or the more detailed KBs as pathway~\cite{pathguide} or Gene Ontology Annotation \cite{goa}. Community challenges help curate these KBs via text mining at a large-scale. E.g., 
BioCreative for PPI~\cite{biocreativeIIppi,biocreativeiiippi}, protein-mutation associations~\cite{mutations}, and gene-disease relations~\cite{genedisease}; or BioNLP~\cite{bionlp09} for complex n-ary bio events. CS NER is also been addressed in equivalent series such as SemEval~\citeyearpar{scienceie,gabor-etal-2018-semeval,ncg} which is promising to foster rapid task progress.

\subsection{Chemistry NER (ChemNER)}

BioNER in part fosters Chemistry NER. Text mining for drug and chemical compound entities~\cite{biocreativeDDI,chemdnerCorpus} are indispensable to mining chemical disease relations \cite{biocreativeVtaskcorpus}, and drug and chemical-protein interactions \cite{biocreativeVIIchemprot,drugprot}. Obtaining this structured knowledge has implications in precision medicine, drug discovery as well as basic biomedical research. Corpora for ChemNER are \citeauthor{chemnerdata}'s dataset (42 full-text papers with \texttildelow7000 chemical entities), ChemDNER (10,000 PubMed abstracts with 84,355 chemical entities) \citeyearpar{chemdnerCorpus}, and NLM-Chem (150 full-text papers with 38,342 chemical entities normalized to 2,064 MeSH identifiers) \cite{nlmchem}.

The high-level identification of entities in text is a staple of most modern NLP pipelines over commonsense knowledge. This, in the context of the scientific entities formalisms presented above is pertinent to scholarly knowledge as well. While being structurally domain-wise related, our work has a unique objective: \textit{to extract entities across STEM Science which follow a generic four-entity conceptual formalism.} Thus apart from having an impact in the emerging field of the discovery of science graphs, the STEM-NER-60k corpus can have specific applications in higher-level NLP tasks including information extraction~\cite{shah2003information}, factuality ascertainment of statements in knowledge base population~\cite{adel2018deep}, and question answering over linked data~\cite{unger2014question}.

\section{Our STEM-NER-60k Corpus}

We now describe our corpus: 1) definitions of the four scientific concepts~\citeyearpar{stem-ecr}, viz. \textsc{process}, \textsc{method}, \textsc{material}, and \textsc{data}, are given; 2) the process by which the silver-standard STEM-NER-60k corpus is created is explained; and 3) corpus insights are offered specifically in terms of the multidisciplinary entities annotated under the formalism of four concepts bridging the 10 domains.


\subsection{Concept Definitions}
\label{four-concepts}

Following an iterative process of concept refinement~\cite{stem-ecr} via a process that involved expert adjudication of which scientific concepts were multidisciplinarily semantically meaningful, the following four concepts were agreed upon to be relevant for entities across STEM, specifically across the following 10 domains, viz. \textit{agri}, \textit{ast}, \textit{bio}, \textit{chem}, \textit{cs}, \textit{es}, \textit{eng}, \textit{ms}, and \textit{math}.

\begin{itemize}
  \item \textbf{\textsc{process}}. Natural phenomenon, or independent/dependent activities. E.g., growing (\textit{bio}), cured (\textit{ms}), flooding (\textit{es}).
  \item \textbf{\textsc{method}}. A commonly used procedure that acts on entities. E.g., powder X-ray (\textit{chem}), the PRAM analysis (\textit{cs}), magnetoencephalography (\textit{med}).
  \item \textbf{\textsc{material}}. A physical or digital entity used for scientific experiments. E.g., soil (\textit{agri}), the moon (\textit{ast}), the set (\textit{math}).
  \item \textbf{\textsc{data}}. The data themselves, or quantitative or qualitative characteristics of entities. E.g., rotational energy (\textit{eng}), tensile strength (\textit{ms}), vascular risk (\textit{med}).
\end{itemize}

\subsection{Corpus Creation}


The silver-standard STEM-NER-60k corpus was created as follows. Roughly 60,000 articles in text format and restricted only to the articles with the CC-BY redistributable license on Elsevier were first downloaded \url{https://tinyurl.com/60k-raw-dataset}. Next, our aim was to obtain the four-concept entity annotations for the Abstracts in this corpus of publications. We leveraged our prior-developed state-of-the-art NER system for this purpose~\cite{brack2020domain}. The \citeauthor{brack2020domain} system was based on \citeauthor{beltagy2019scibert}’s SciBERT which in turn for the specific NER configuration makes use of the original BERT NER~\cite{devlin,Ma2016EndtoendSL} prediction architecture. The ~\citeauthor{brack2020domain} system, however, was pretrained on a smaller gold-standard STEM corpus~\citeyearpar{stem-ecr} expert-annotated with the four generic scientific entities described in \autoref{four-concepts}. Applying this system on the Abstracts on the newly downloaded 60k articles produced the silver-standard STEM-NER-60k corpus of this work. The resulting silver-standard corpus statistics are shown in \autoref{table:stats}.



\begin{table}[!htb]
\begin{tabular}{p{0.65cm}|p{0.9cm}|p{0.9cm}|p{0.9cm}|p{1.1cm}|p{1.0cm}}
  & articles & process & method & material & data \\ \hline
\textit{agr}     & 4,944      & 20,532  & 3,252  & 62,043   & 33,608  \\
\textit{ast}     & 15,003     & 31,104  & 10,423 & 55,753   & 97,011  \\
\textit{bio}     & 9,038      & 54,029  & 6,777  & 100,454  & 43,418  \\
\textit{chem}    & 5,232      & 18,185  & 6,044  & 48,779   & 30,596  \\
\textit{cs}      & 5,389      & 17,014  & 13,650 & 35,554   & 33,575  \\
\textit{es}      & 4,363      & 28,432  & 5,129  & 56,571   & 50,211  \\
\textit{eng}     & 2,441      & 12,705  & 3,293  & 24,633   & 24,238  \\
\textit{ms}      & 2,144      & 10,102  & 2,437  & 23,054   & 16,981  \\
\textit{math}    & 1,765      & 8,002   & 1,941  & 11,381   & 15,631  \\
\textit{med}     & 15,054     & 89,637  & 19,580 & 148,059  & 134,249
\end{tabular}
\caption{STEM-NER-60k corpus statistics comprising 59,984 articles structured in terms of \textsc{process}, \textsc{method}, \textsc{material}, and \textsc{data} concepts}
\label{table:stats}
\end{table}

Our corpus is publicly released (\url{https://github.com/jd-coderepos/stem-ner-60k}) to support related R\&D endeavors and for other researchers interested in further investigating the task of scholarly NER. 

\subsection{Corpus Insights}

Here, details of the STEM-NER-60k corpus are offered.

But first, given our large-scale multidisciplinary scientific corpus, we take the opportunity to briefly examine the difference between scientific writing and non-scientific writing, if any, with the help of entropy formulations for each of our data domains. The entropies obtained for our corpus per domain is as follows: \textit{med} (4.58) > \textit{chem} (4.58) > \textit{bio} (4.56) > \textit{ast} (4.53) > \textit{agri} (4.5) > \textit{ms} (4.5) > \textit{es} (4.48) > \textit{eng} (4.42) > \textit{math} (4.4). Intriguingly, these numbers are close to non-scientific English (4.11 bits~\cite{cover}). Thus, given our corpus scientific English, one can rule out any atypical English usage syntax other than domain-specific jargon vocabulary.

Next, briefly, we address \textit{which general scientific entity annotation patterns can one anticipate in our silver-standard corpus?} 1) Entity annotations can be expected as definite noun phrases whenever found. 2) Coreferring lexical units for scientific entities in the context of a single Abstract can be expected to be annotated with the same concept type. 3) Quantifiable lexical units such as numbers (e.g., years 1999, measurements 4km) or even phrases (e.g., vascular risk) should be \textsc{data}. 4) Where found, the most precise text reference (i.e. including qualifiers) regarding \textsc{material}s used in the experiment should be annotated. E.g., the term ``carbon atoms in graphene'' was annotated as a single \textsc{material} entity and not separately as ``carbon atoms'', ``graphene''. 5) The precedence of annotation of the scientific concepts in the original corpus in case of any confusion in classifying the four classes of scientific entities was resolved as follows: \textsc{method} > \textsc{process} > \textsc{data} > \textsc{material}, where the concept appearing earlier in the list was selected as the preferred class.

\subsubsection{STEM scientific terms as \textsc{process}}

Verbs (e.g., measured), verb phrases (e.g., integrating results), or noun phrases (e.g., an assessment, future changes, this transport process, the transfer) can be expected to be scientific entity candidates for \textsc{process}. Generally, it can be one of two things: an occurrence natural to the state/properties of the entity or an action performed by the investigators. In the latter case, however, it is better aptly expected as a \textsc{method} entity when the action is a named instance.

Some examples are offered for scientific entities as \textsc{process} candidates. 1) In the sentence, ``The transfer of a laboratory process into a manufacturing facility is one of the most critical steps required for the large scale production of cell-based therapy products'', the terms “The transfer”, “a laboratory process”, and “the large scale production” all are of type \textsc{process}. 2) In ``The transterminator ion flow in the Venusian ionosphere is observed at solar minimum for the first time.'', the terms ``The transterminator ion flow'' and ``solar minimum'' \textsc{process} entities. The verb ``observed'', however, is not considered a \textsc{process} since it doesn’t act upon another object. 3) On the other hand, in ``It is suggested that this ion flow contributes to maintaining the nightside ionosphere.'', the terms ``this ion flow'' and ``maintaining'' are both considered valid \textsc{process} candidates. Finally, 4) in the sentence ``Cellular morphology, pluripotency gene expression and differentiation into the three germ layers have been used compare the outcomes of manual and automated processes'' the terms ``pluripotency gene expression'', ``differentiation'', ``compare'', and ``manual and automated processes'' are each annotated as \textsc{process}.

\subsubsection{STEM scientific terms as \textsc{method}}

Phrases which can be expected to be annotated as a \textsc{method} entity are those which contain the following trigger words: simulation, method, algorithm, scheme, technique, system, function, derivative, proportion, strategy, solver, experiment, test, computation, program. As an example, consider the sentence ``Here finite-element modelling has demonstrated that once one silica nanoparticle debonds then its nearest neighbours are shielded from the applied stress field, and hence may not debond.'' In this sentence, the term ``finite-element modelling'' is annotated as a \textsc{method}.

\subsubsection{STEM scientific terms as \textsc{material}}

This concept is expounded merely with the following three examples. 1) In ``Based on the results of the LUCAS topsoil survey we performed an assessment of plant available P status of European croplands.'' the term ``European croplands'' should be \textsc{material}. 2) In ``The transfer of a laboratory process into a manufacturing facility is one of the most critical steps required for the large scale production of cell-based therapy products.'' there are two \textsc{material} terms, viz. ``a manufacturing facility'' and ``cell-based therapy products''. Finally, 3) in the sentence ``Cellular morphology, pluripotency gene expression and differentiation into the three germ layers have been used to compare the outcomes of manual and automated processes.'' the phrase ``the three germ layers'' is a \textsc{material}.

\subsubsection{STEM scientific terms as \textsc{data}}

Phrases satisfying the patterns in the following examples can be expected to be \textsc{data}. 1) In ``Based on the results of the LUCAS topsoil survey we performed an assessment of plant available P status of European croplands.'', the phrases ``the results'' and ``plant available P status'' are considered as \textsc{data} in the original annotation scheme. 2) In ``Our analysis shows a status of a baseline period of the years 2009 and 2012, while a repeated LUCAS topsoil survey can be a useful tool to monitor future changes of nutrient levels, including P in soils of the EU.'' the phrases: ``a status of a baseline period'', ``nutrient levels'', and ``P'' are \textsc{data} items. 3) Further, in ``Observations near the terminator of the energies of ions of ionospheric origin showed asymmetry between the noon and midnight sectors, which indicated an antisunward ion flow with a velocity of (2.5±1.5)kms-1.'' the terms ``asymmetry between the noon and midnight sectors'', ``a velocity'', and ``(2.5±1.5)kms-1'' are \textsc{data}. Finally, 4) in ``We established a P fertilizer need map based on integrating results from the two systems.'' the phrase ``a P fertilizer need map'' is \textsc{data} which should override the selection of ``a P fertilizer'' as \textsc{material} by the concept precedence stated earlier.

Next, in Figures 1 to 10, word clouds of the top 100 entities for the 10 STEM disciplines is shown.

\begin{figure*}[!htb]
    \centering
    \subfloat[\textsc{process}]{{\includegraphics[width=3.9cm]{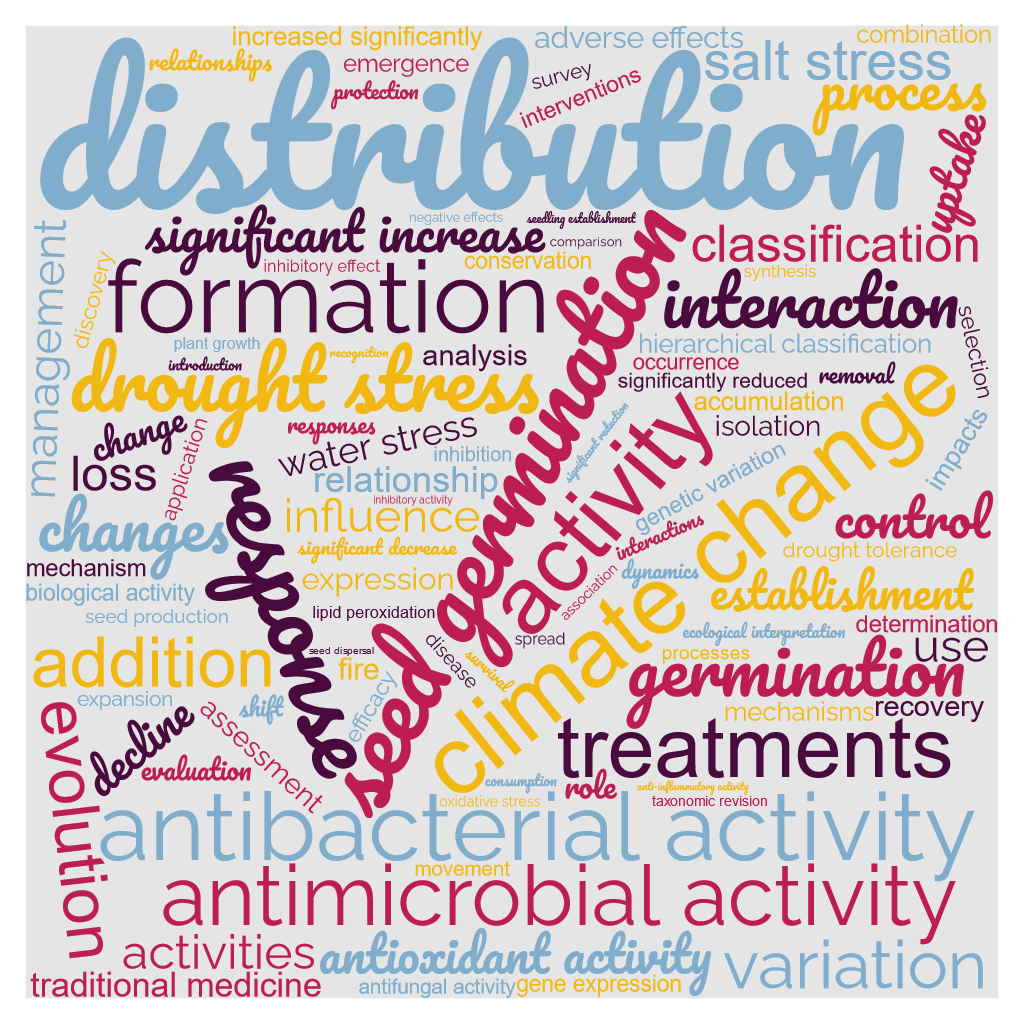} }}
    \subfloat[\textsc{method}]{{\includegraphics[width=3.9cm]{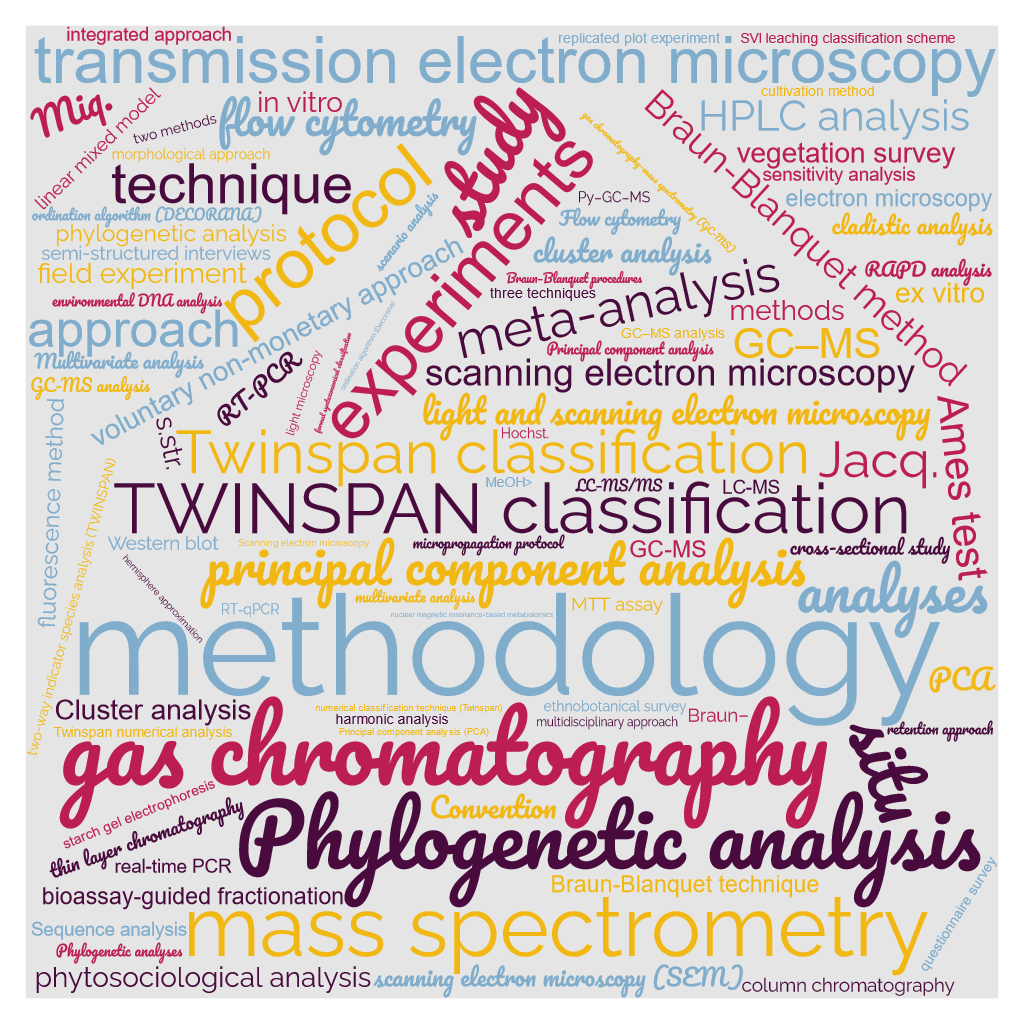} }}
    \subfloat[\textsc{material}]{{\includegraphics[width=3.9cm]{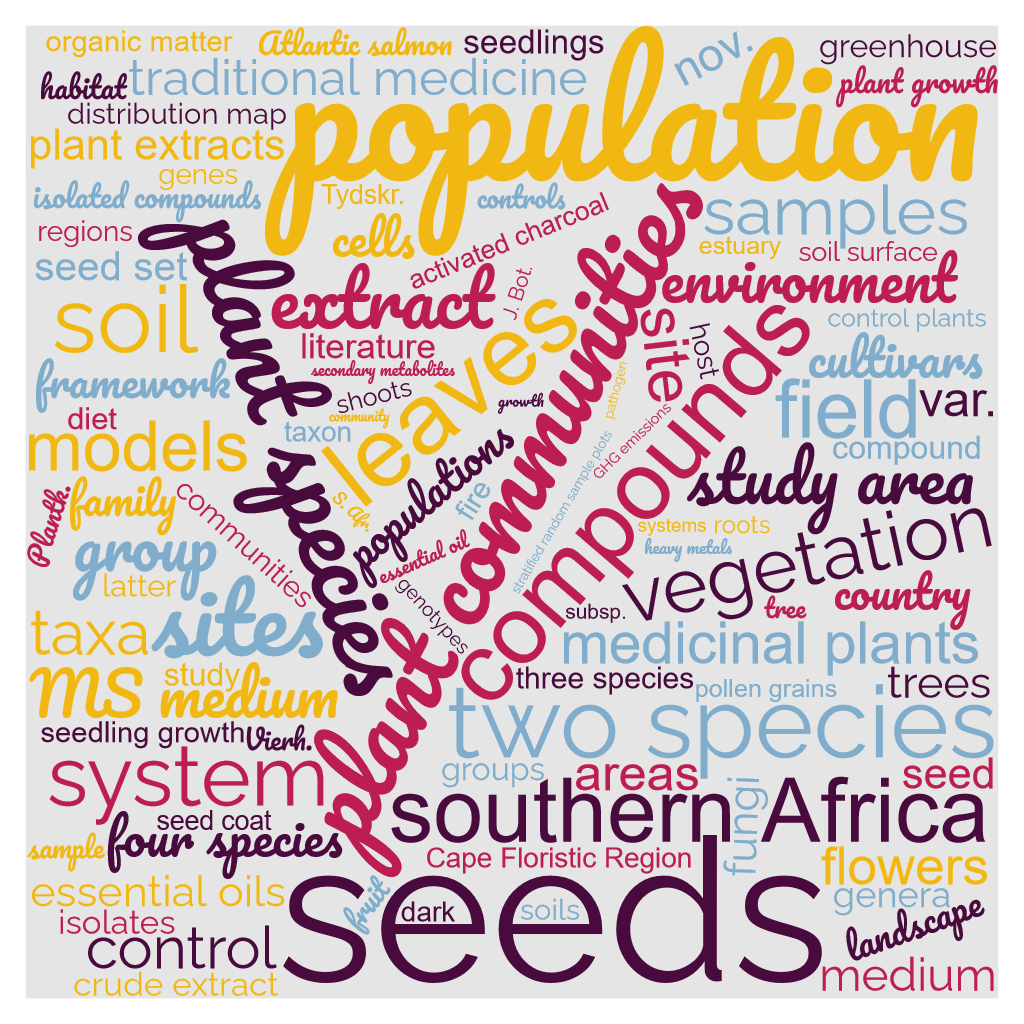} }}
    \subfloat[\textsc{data}]{{\includegraphics[width=3.9cm]{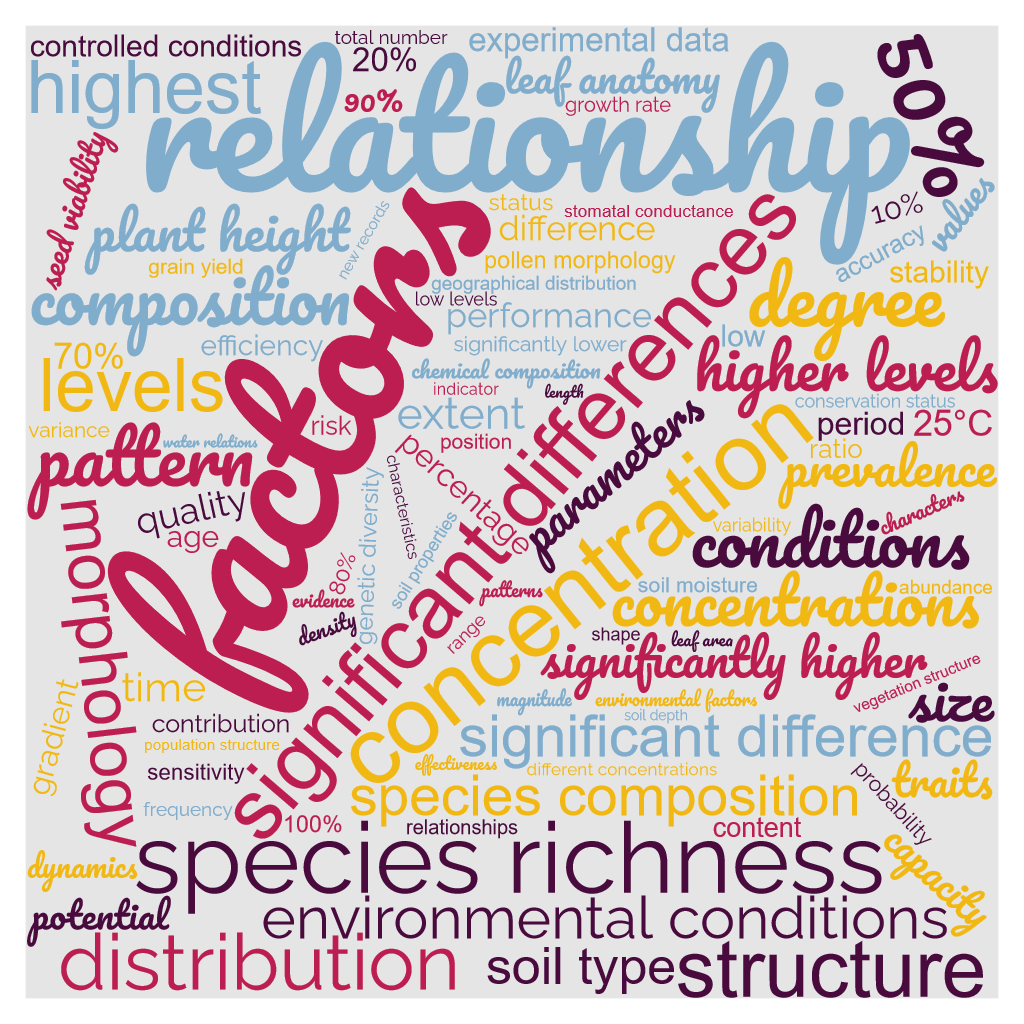} }}    
    \caption{Agriculture domain word clouds}
    \label{fig:agri}
\end{figure*}

\begin{figure*}[!htb]
    \centering
    \subfloat[\textsc{process}]{{\includegraphics[width=3.9cm]{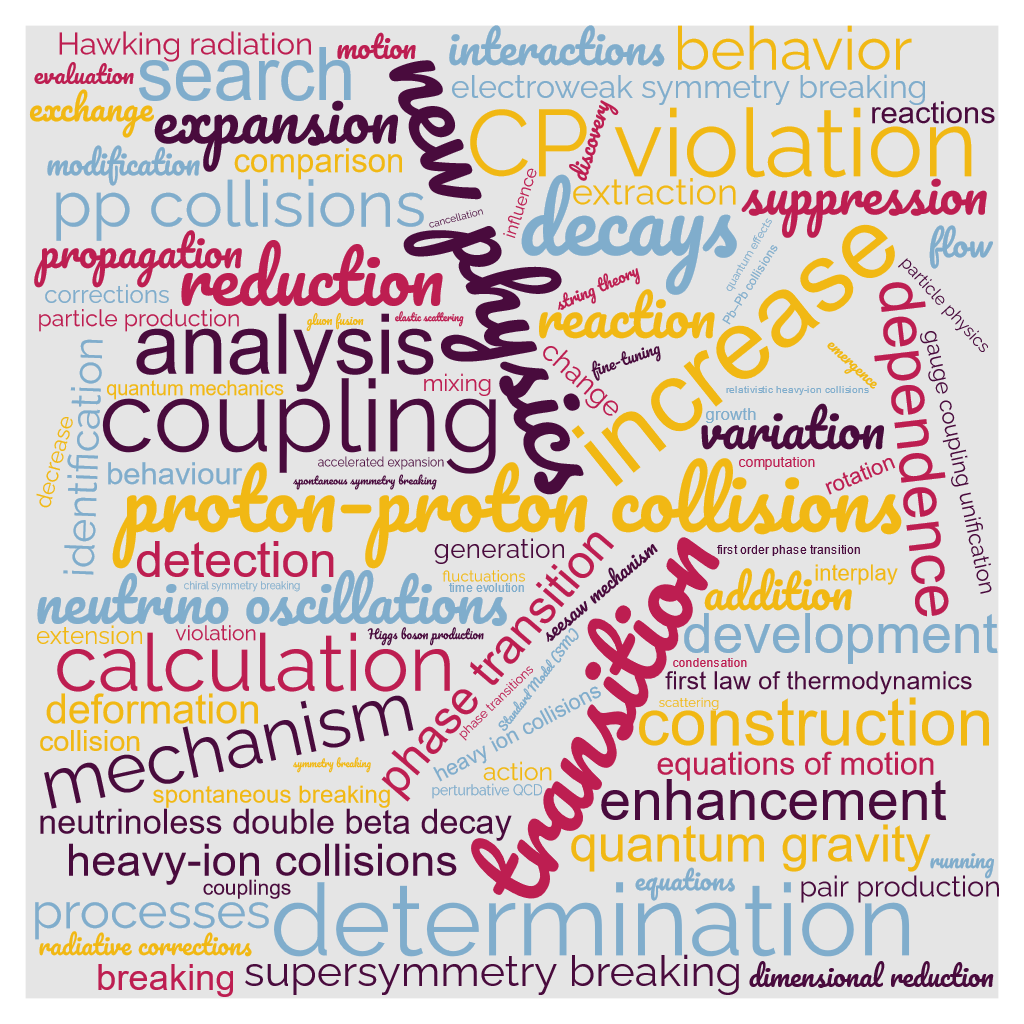} }}
    \subfloat[\textsc{method}]{{\includegraphics[width=3.9cm]{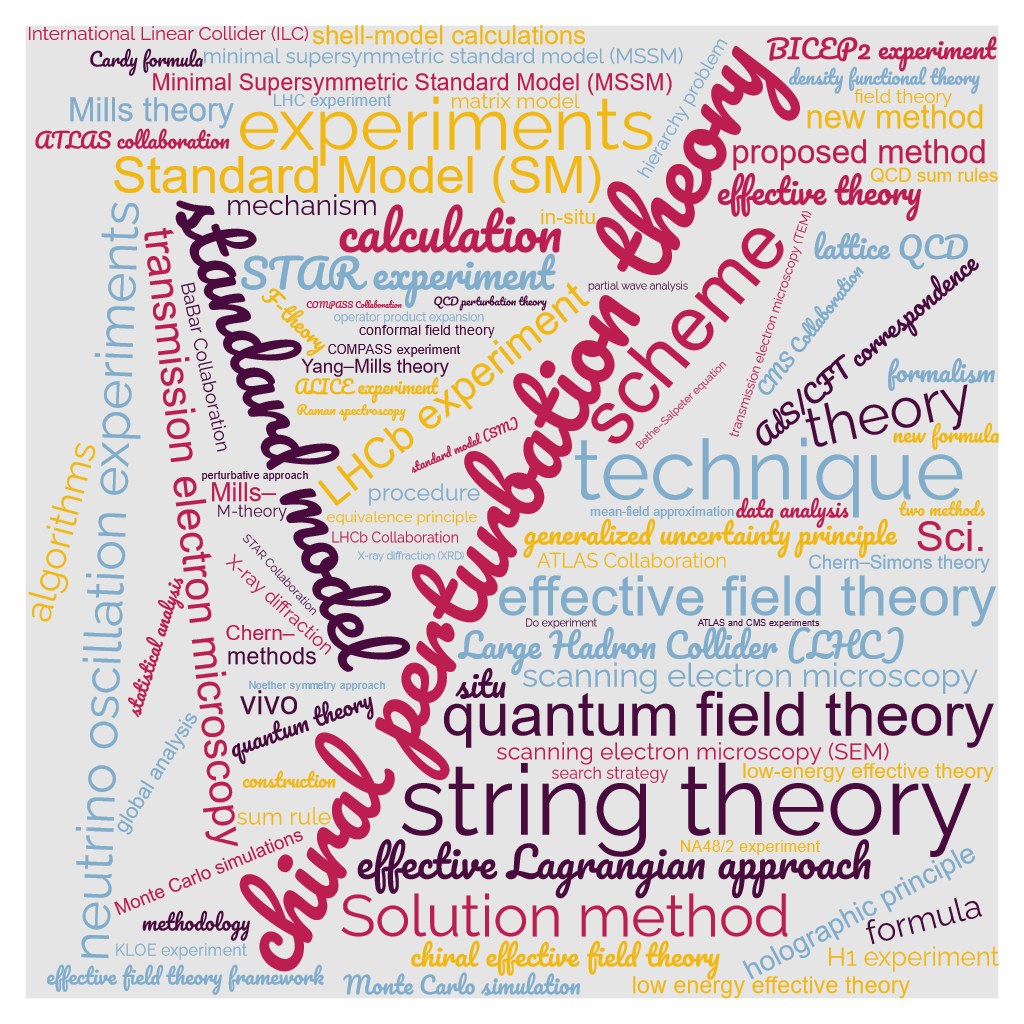} }}
    \subfloat[\textsc{material}]{{\includegraphics[width=3.9cm]{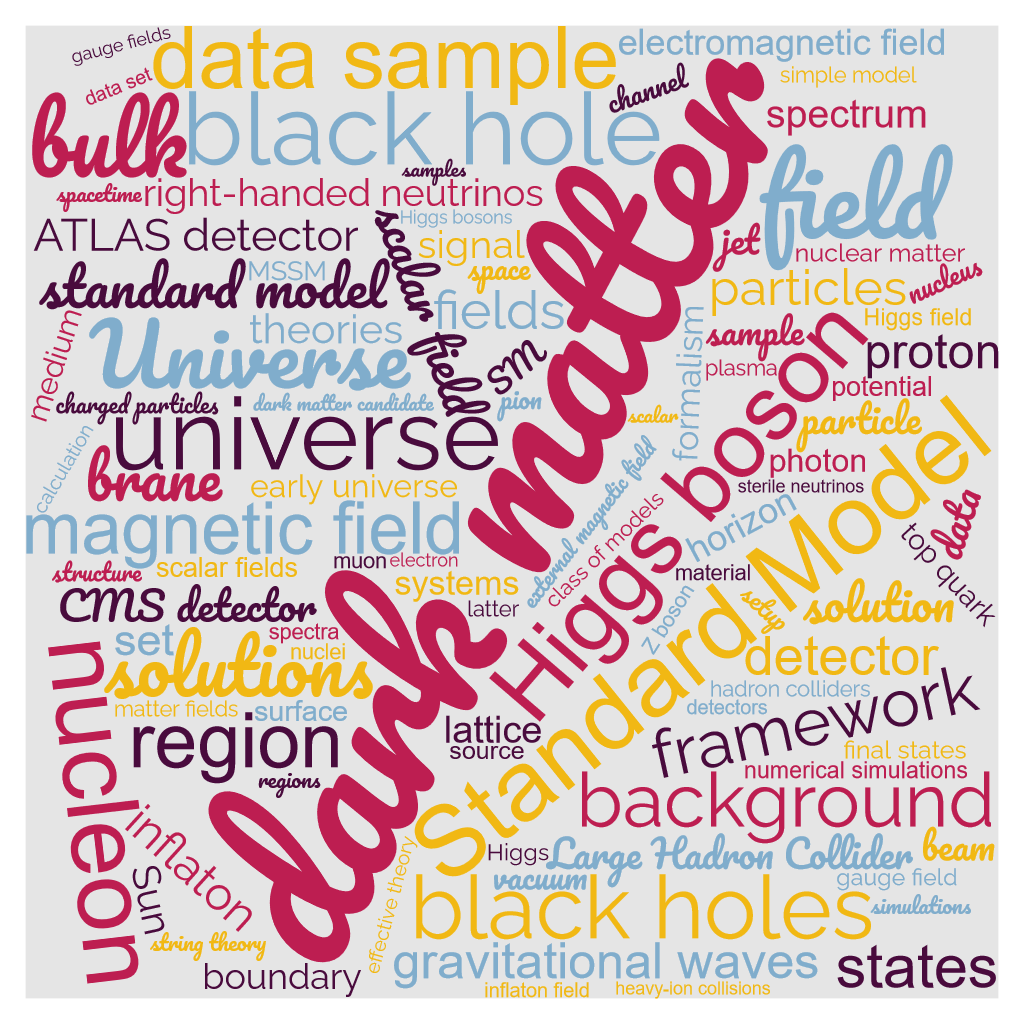} }}
    \subfloat[\textsc{data}]{{\includegraphics[width=3.9cm]{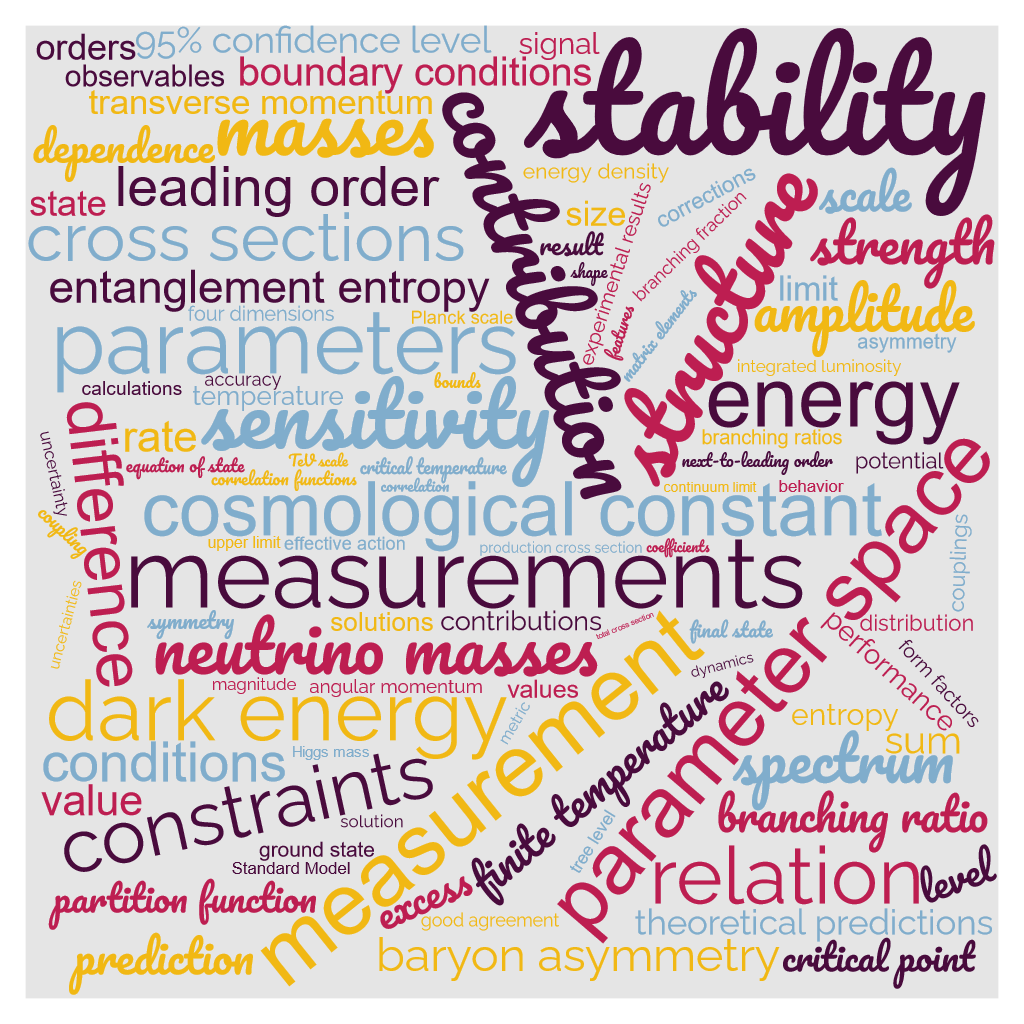} }}    
    \caption{Astronomy domain word clouds}
    \label{fig:ast}
\end{figure*}

\begin{figure*}[!htb]
    \centering
    \subfloat[\textsc{process}]{{\includegraphics[width=3.9cm]{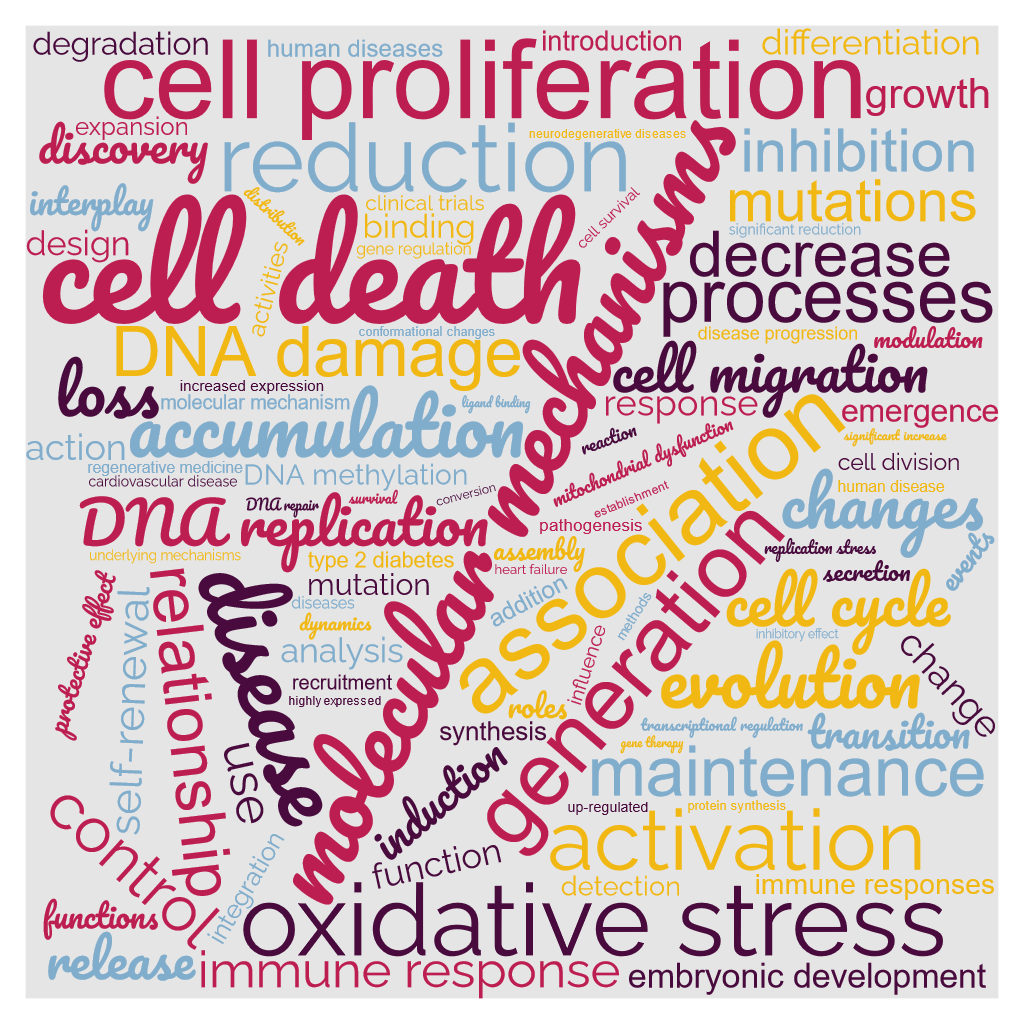} }}
    \subfloat[\textsc{method}]{{\includegraphics[width=3.9cm]{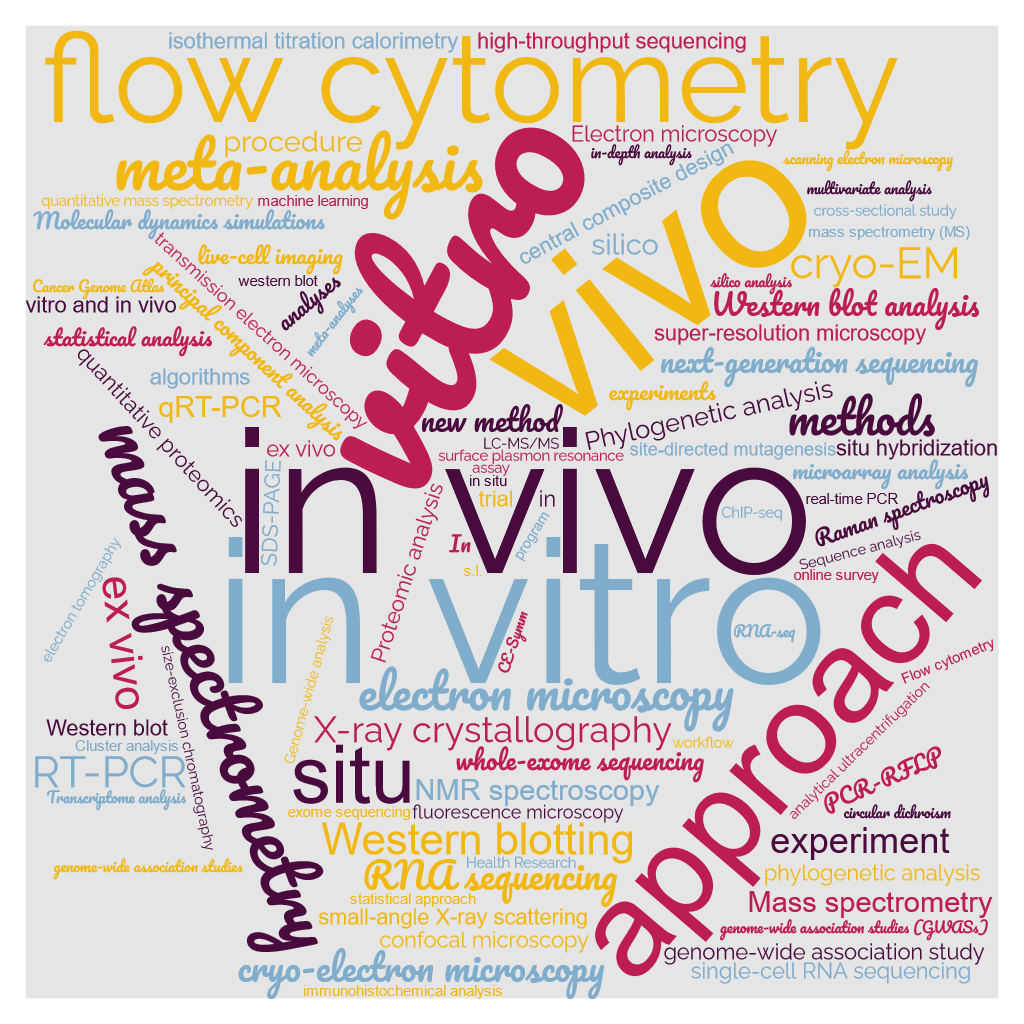} }}
    \subfloat[\textsc{material}]{{\includegraphics[width=3.9cm]{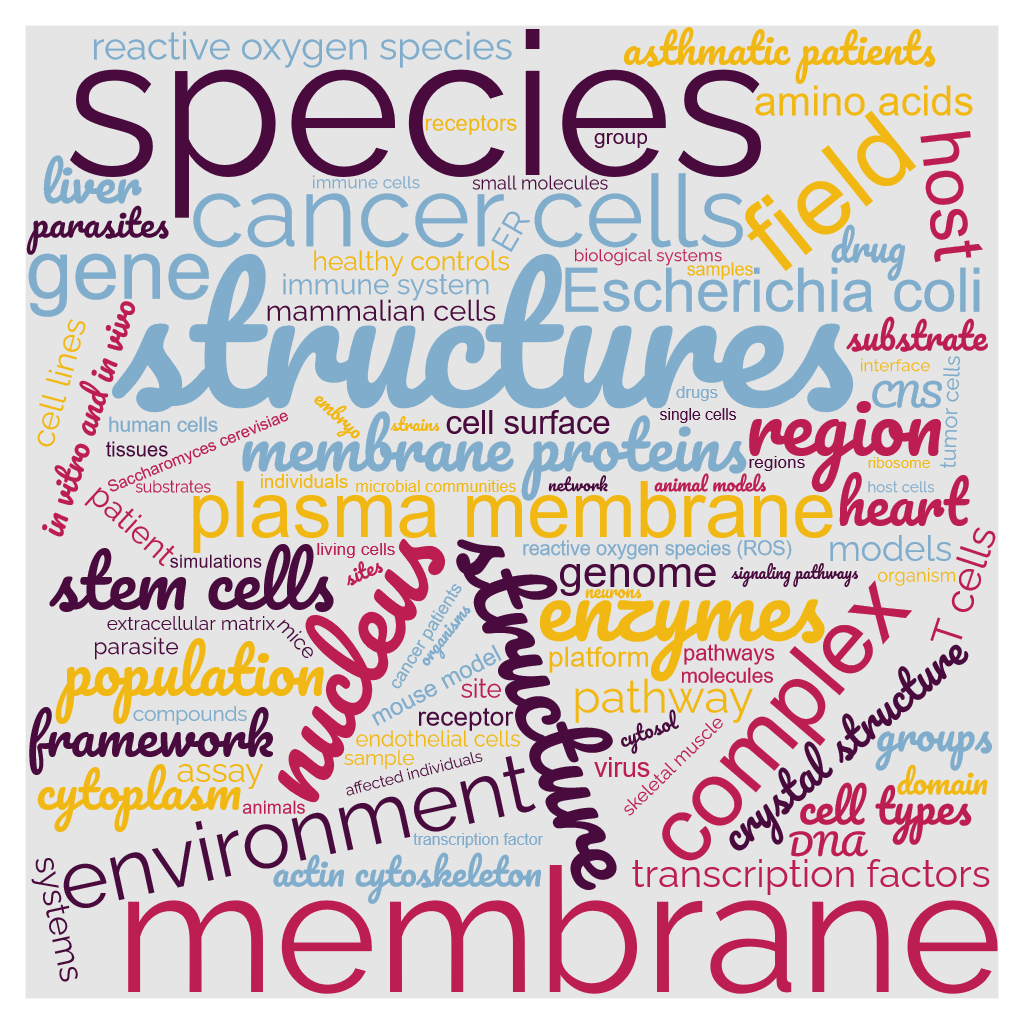} }}
    \subfloat[\textsc{data}]{{\includegraphics[width=3.9cm]{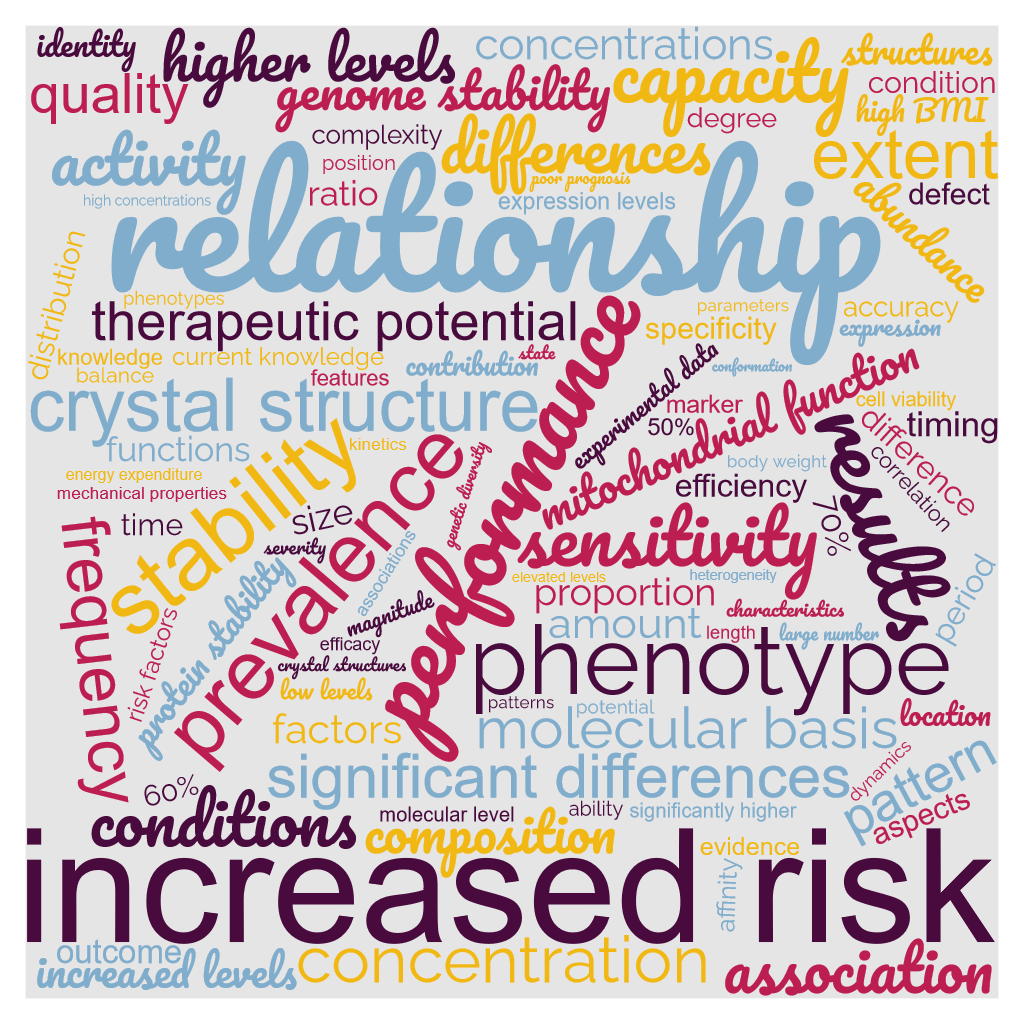} }}    
    \caption{Biology domain word clouds}
    \label{fig:bio}
\end{figure*}

\begin{figure*}[!htb]
    \centering
    \subfloat[\textsc{process}]{{\includegraphics[width=3.9cm]{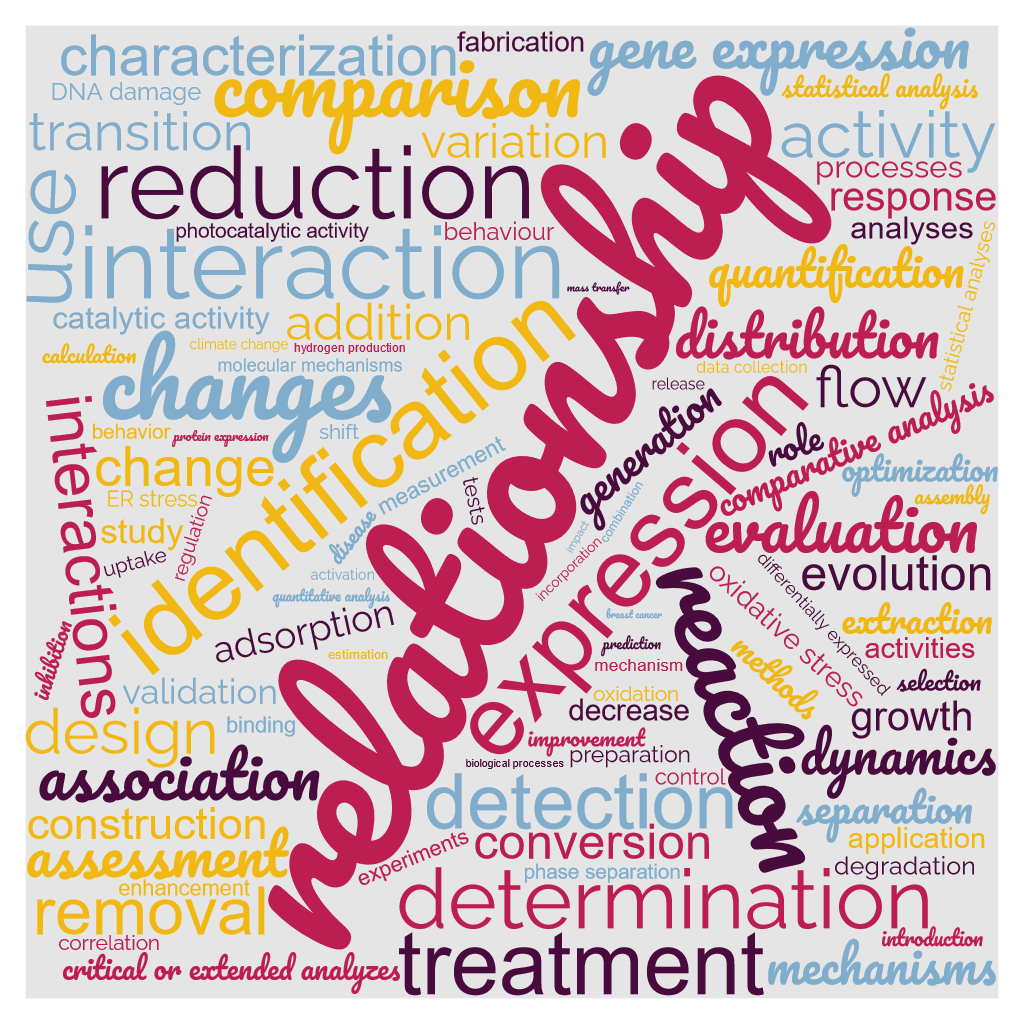} }}
    \subfloat[\textsc{method}]{{\includegraphics[width=3.9cm]{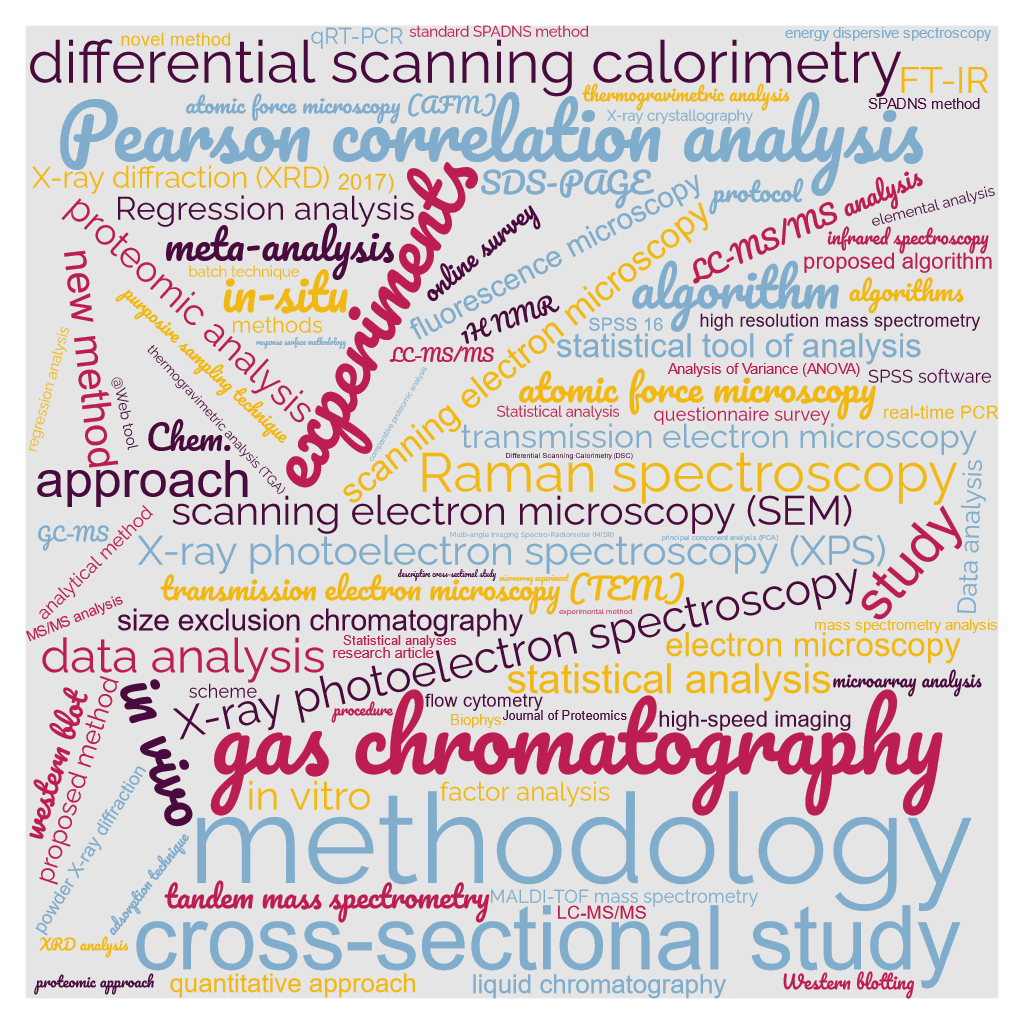} }}
    \subfloat[\textsc{material}]{{\includegraphics[width=3.9cm]{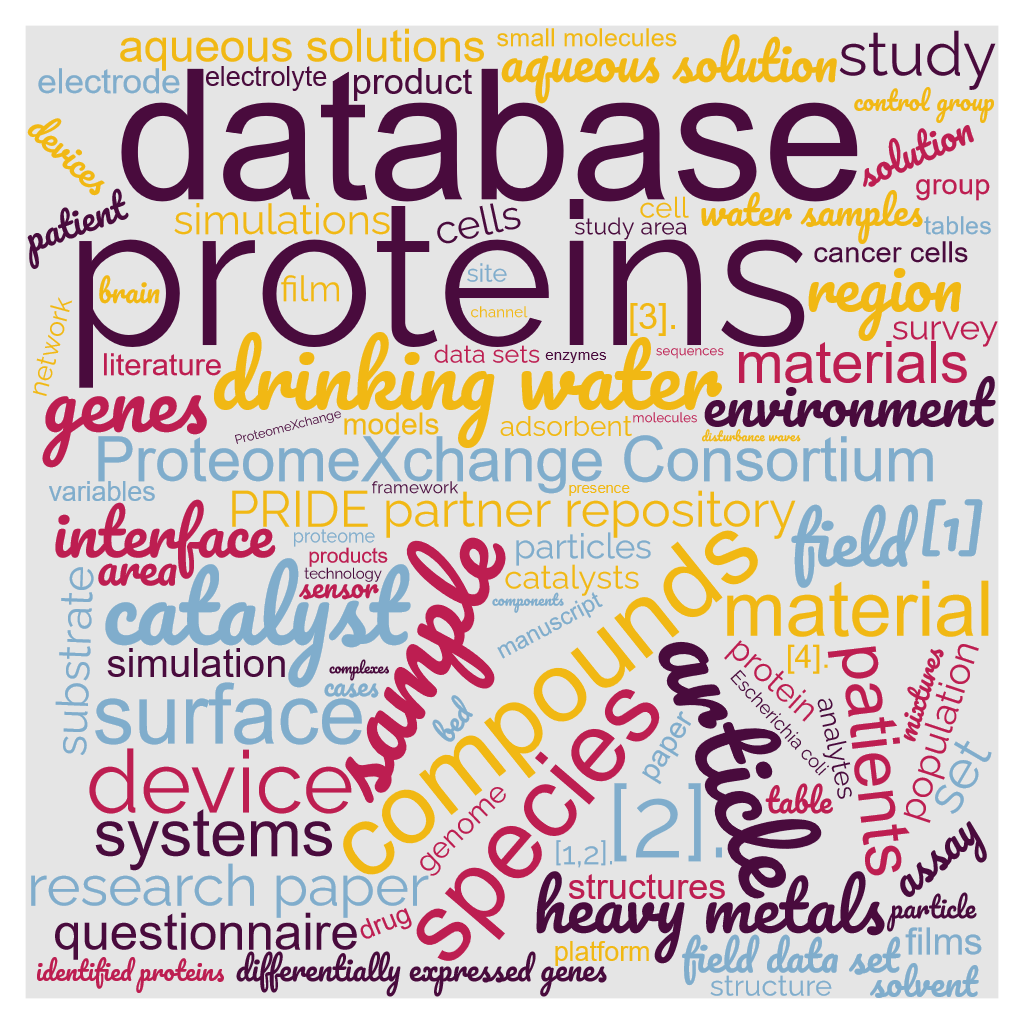} }}
    \subfloat[\textsc{data}]{{\includegraphics[width=3.9cm]{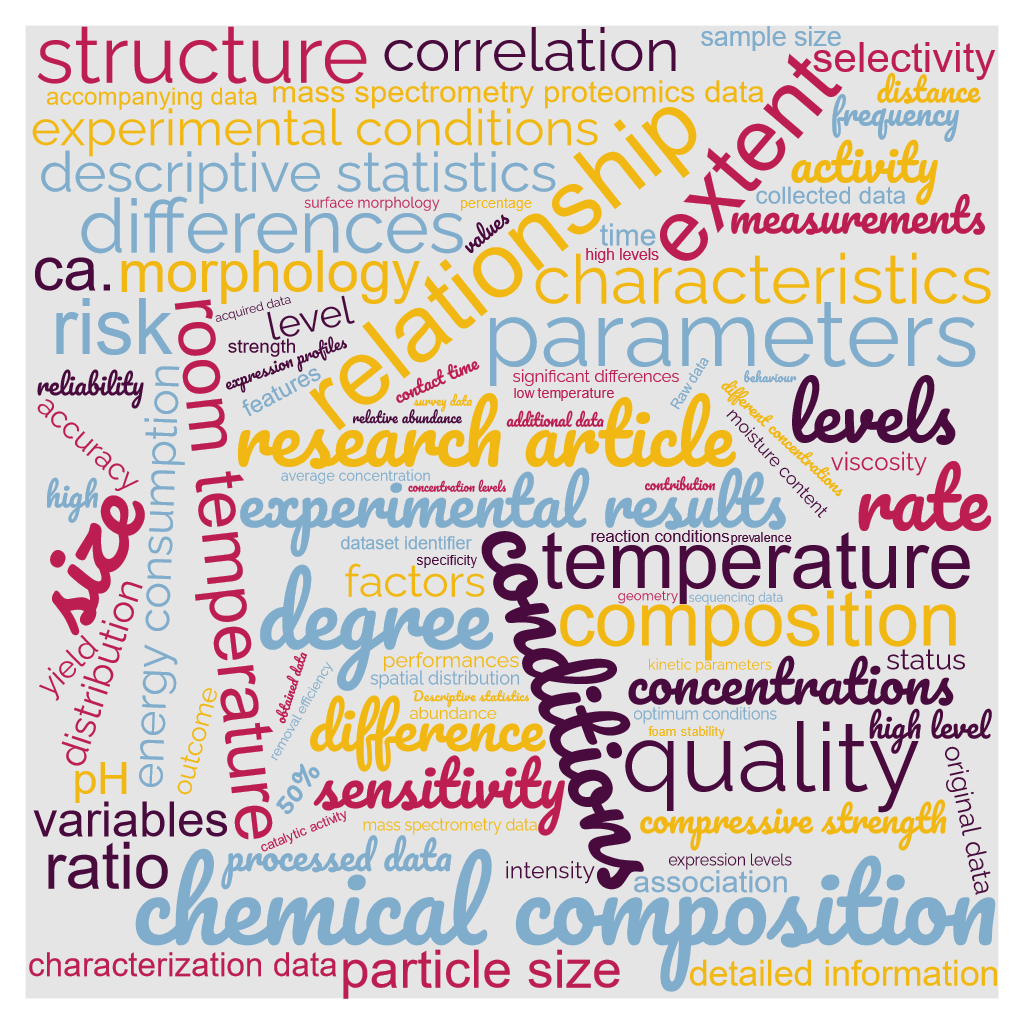} }}    
    \caption{Chemistry domain word clouds}
    \label{fig:chem}
\end{figure*}

\begin{figure*}[!htb]
    \centering
    \subfloat[\textsc{process}]{{\includegraphics[width=3.9cm]{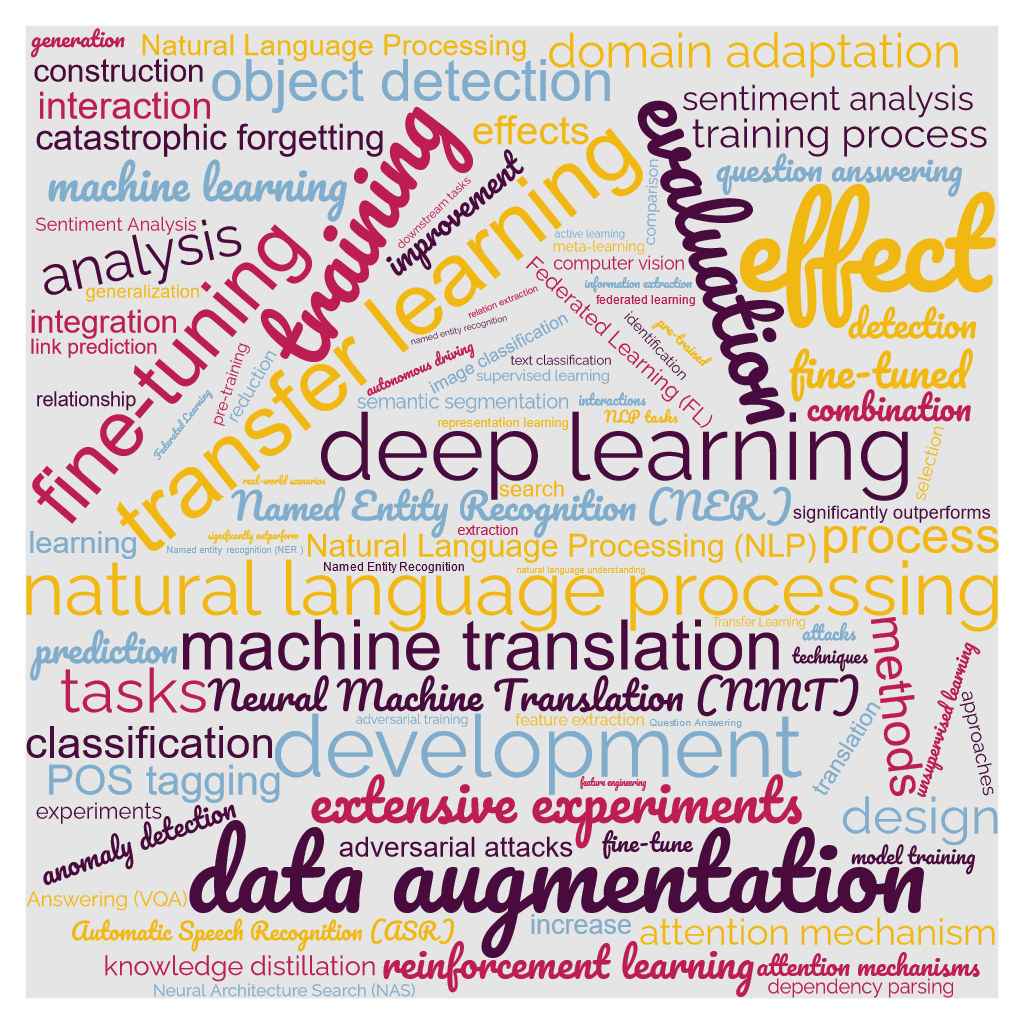} }}
    \subfloat[\textsc{method}]{{\includegraphics[width=3.9cm]{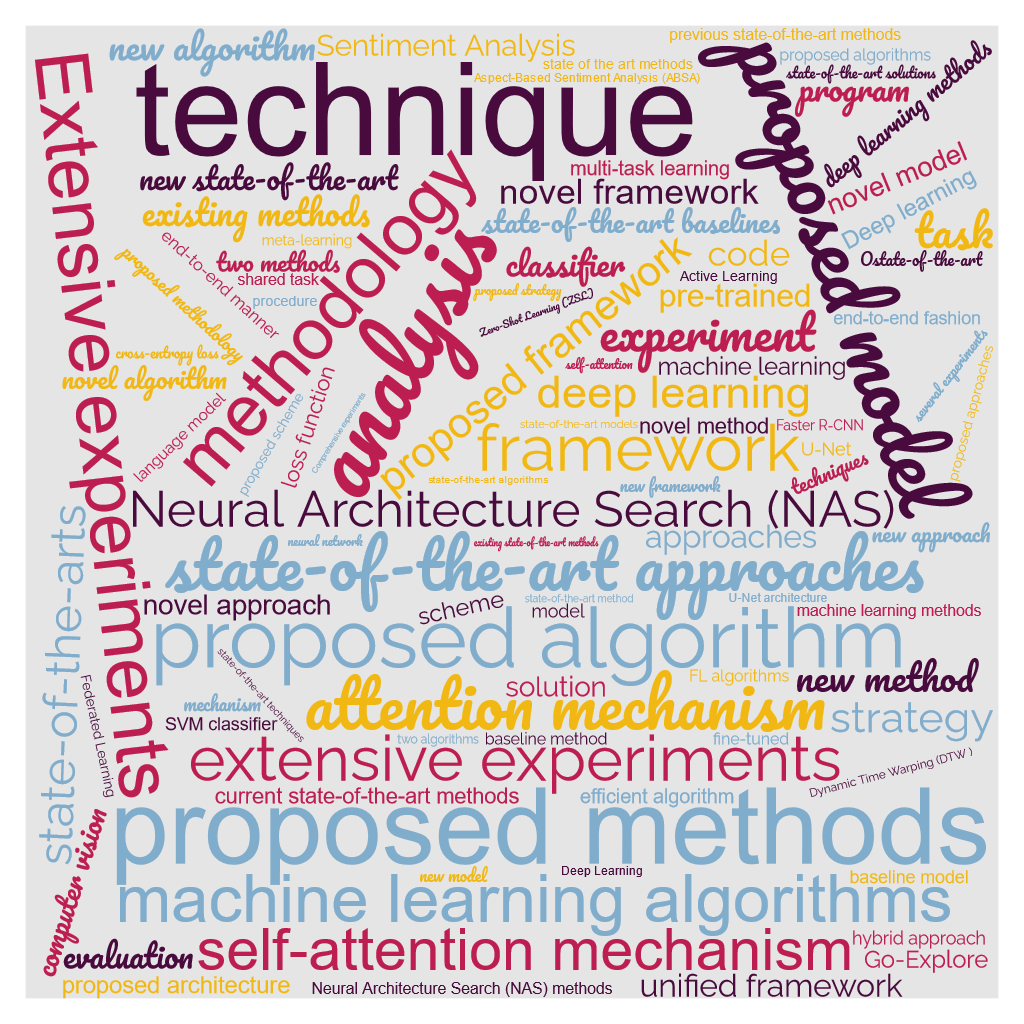} }}
    \subfloat[\textsc{material}]{{\includegraphics[width=3.9cm]{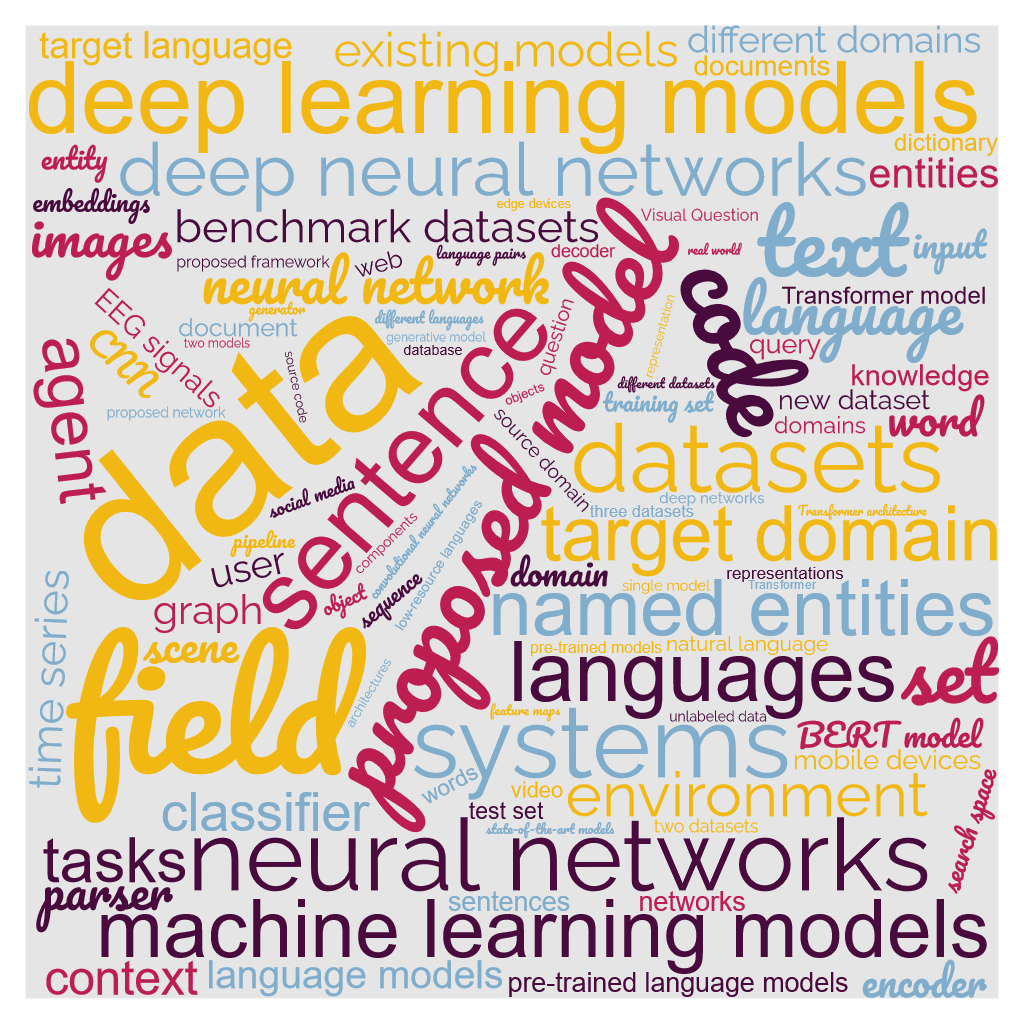} }}
    \subfloat[\textsc{data}]{{\includegraphics[width=3.9cm]{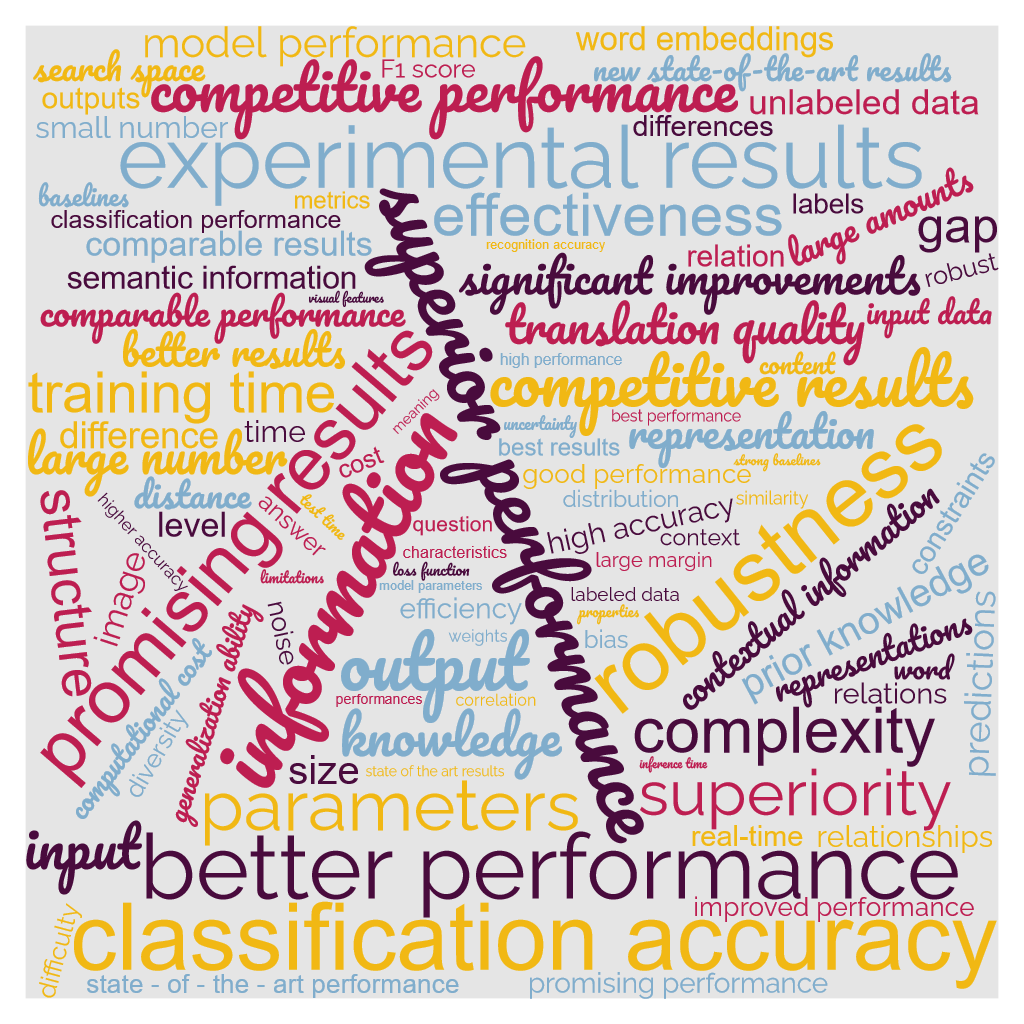} }}    
    \caption{Computer Science domain word clouds}
    \label{fig:cs}
\end{figure*}

\begin{figure*}[!htb]
    \centering
    \subfloat[\textsc{process}]{{\includegraphics[width=3.9cm]{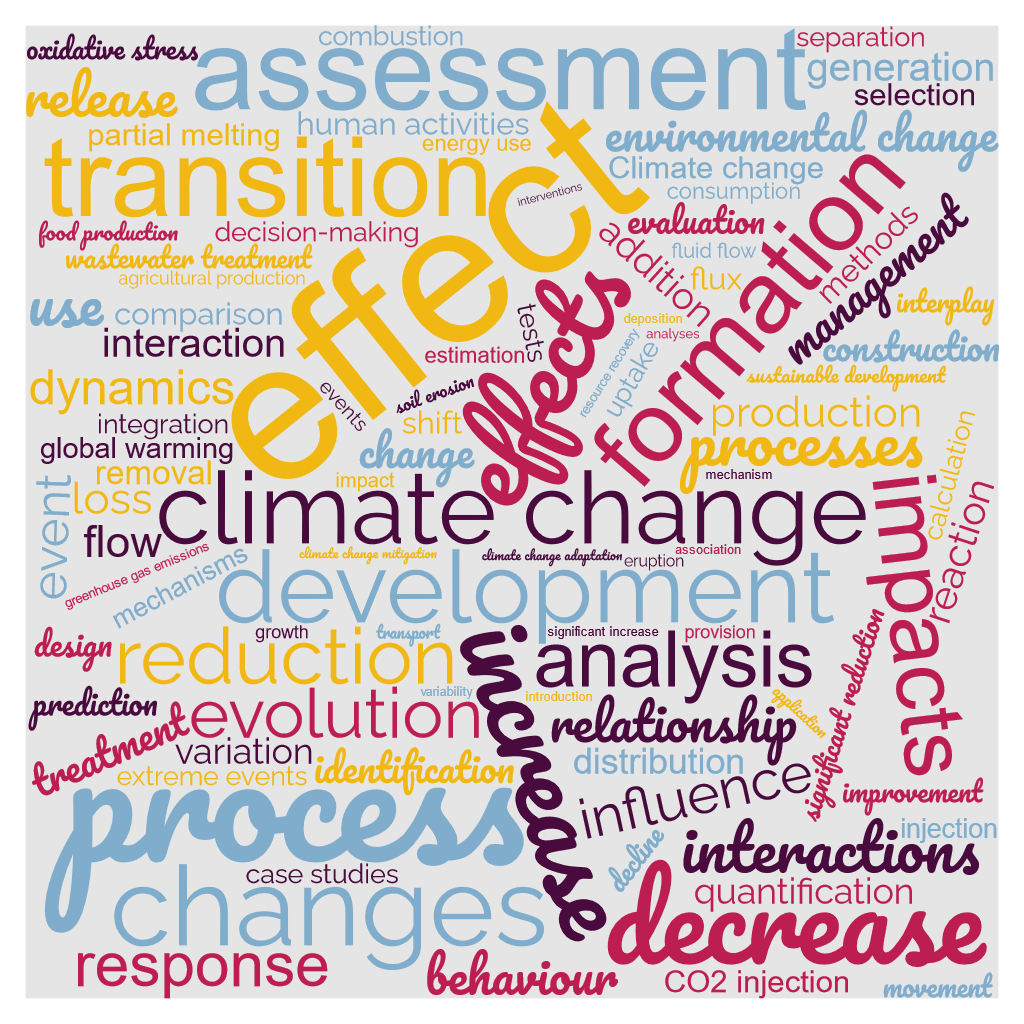} }}
    \subfloat[\textsc{method}]{{\includegraphics[width=3.9cm]{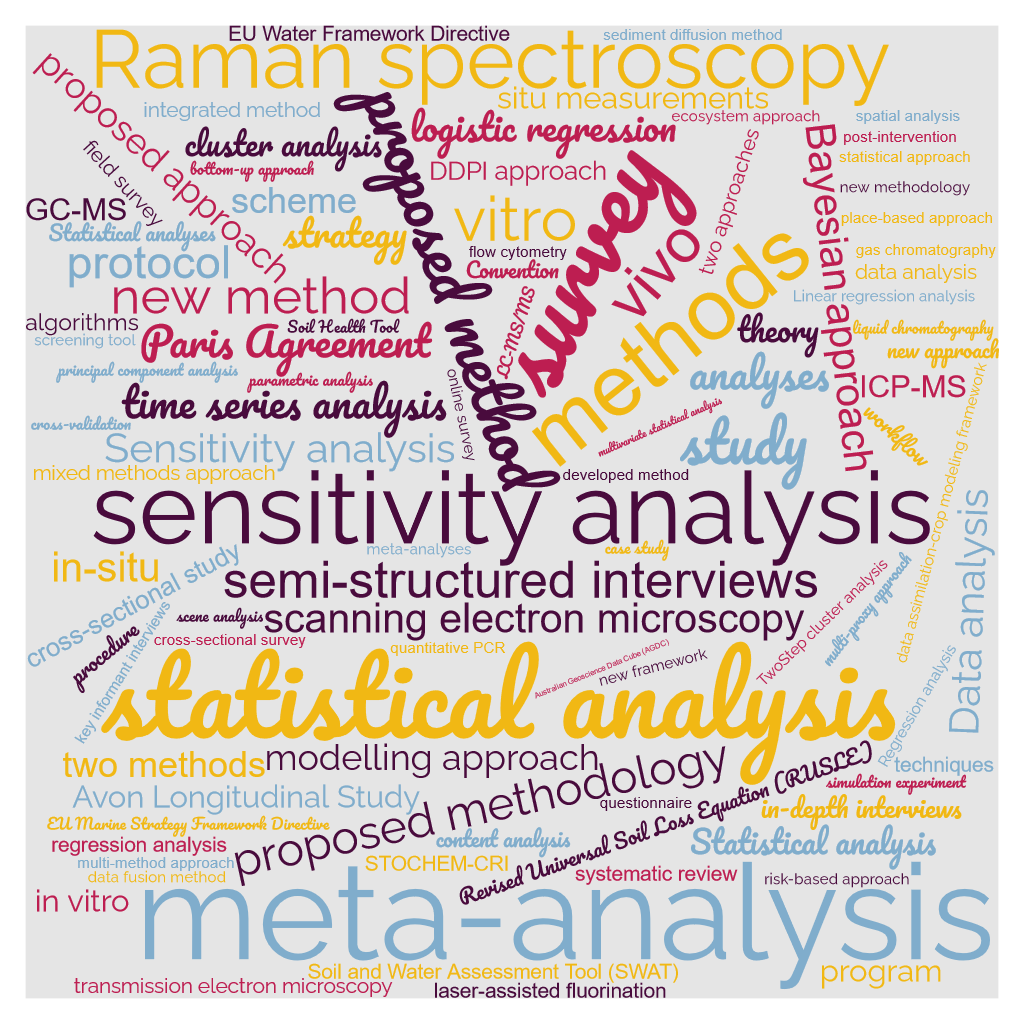} }}
    \subfloat[\textsc{material}]{{\includegraphics[width=3.9cm]{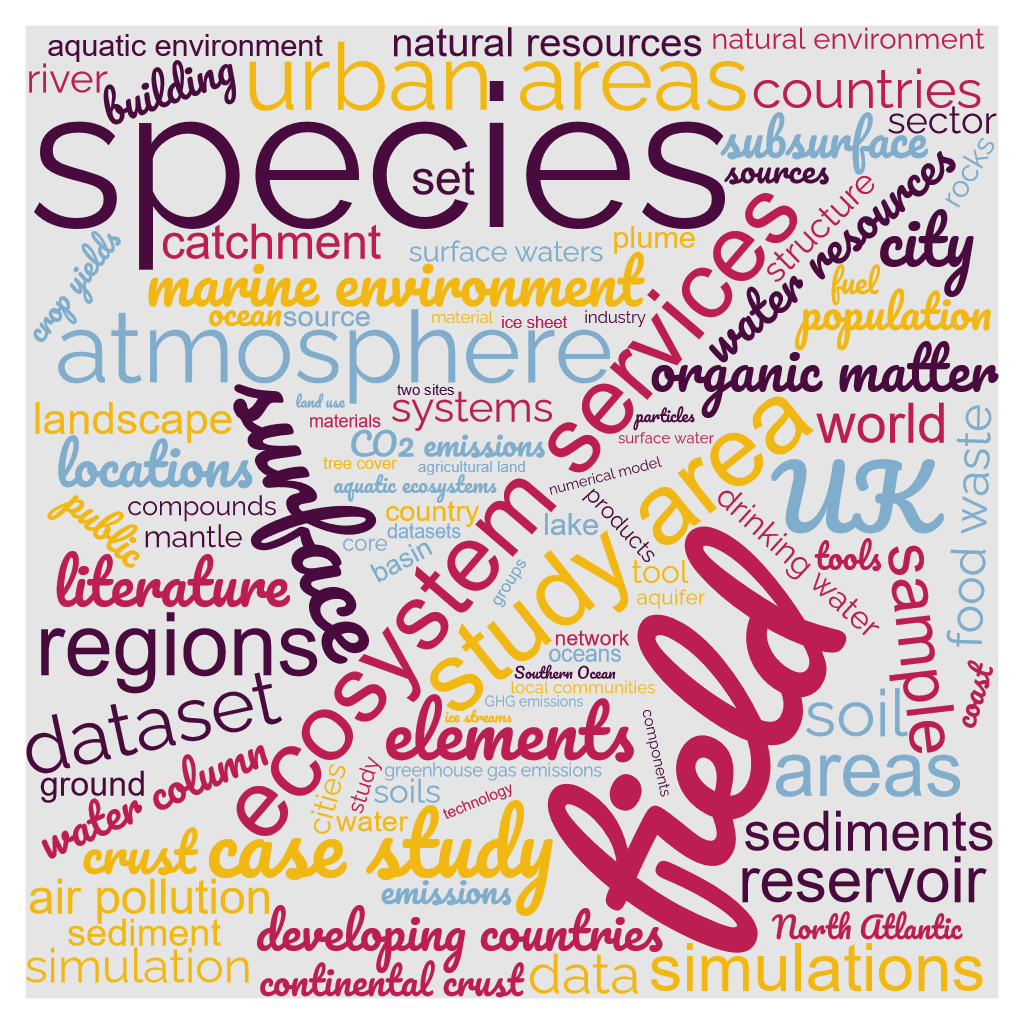} }}
    \subfloat[\textsc{data}]{{\includegraphics[width=3.9cm]{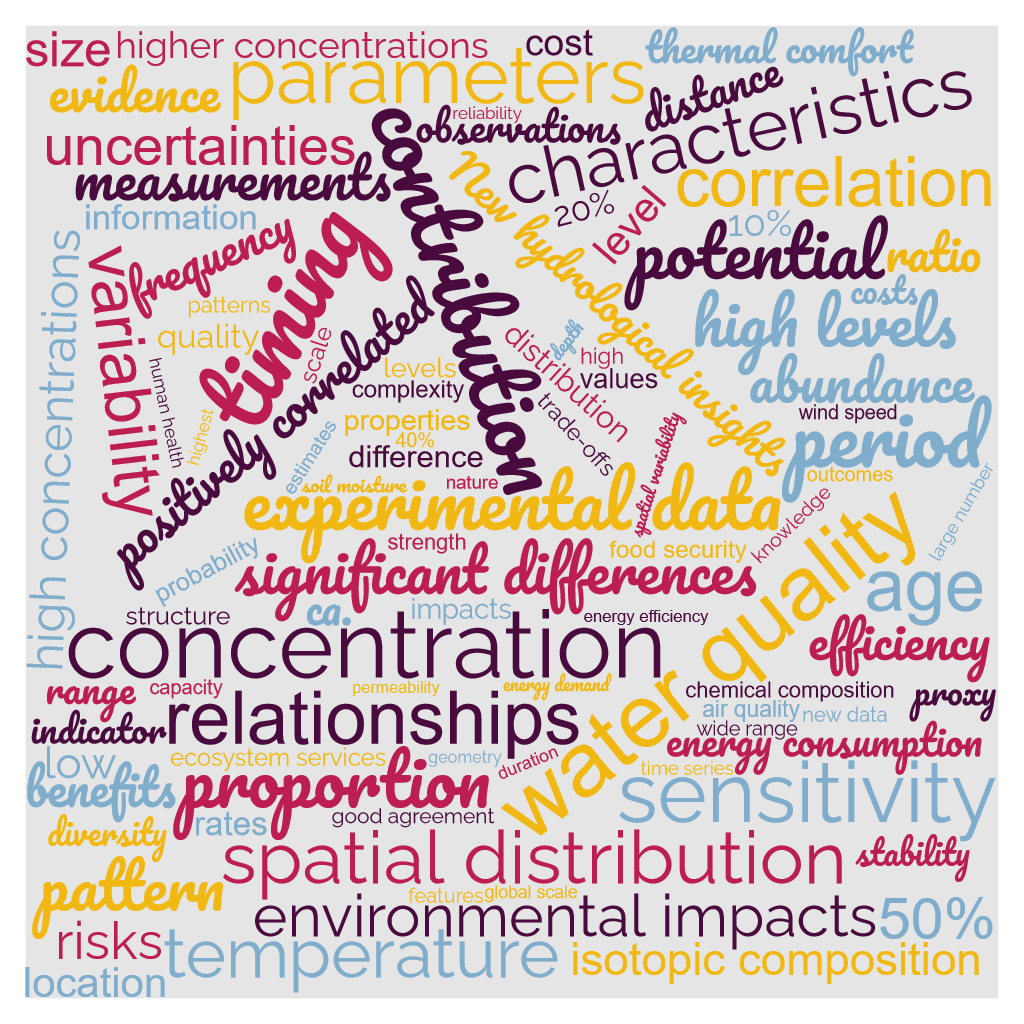} }}    
    \caption{Earth Science domain word clouds}
    \label{fig:es}
\end{figure*}

\begin{figure*}[!htb]
    \centering
    \subfloat[\textsc{process}]{{\includegraphics[width=3.9cm]{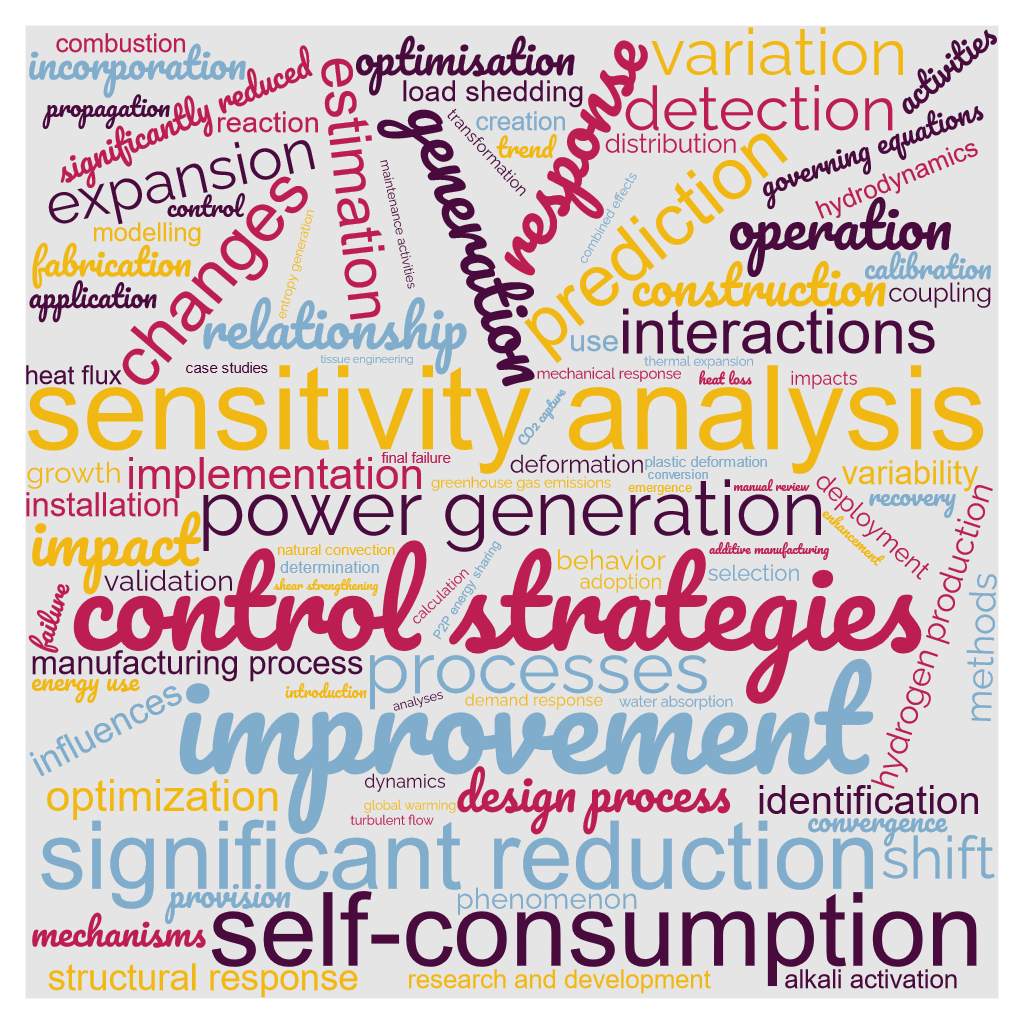} }}
    \subfloat[\textsc{method}]{{\includegraphics[width=3.9cm]{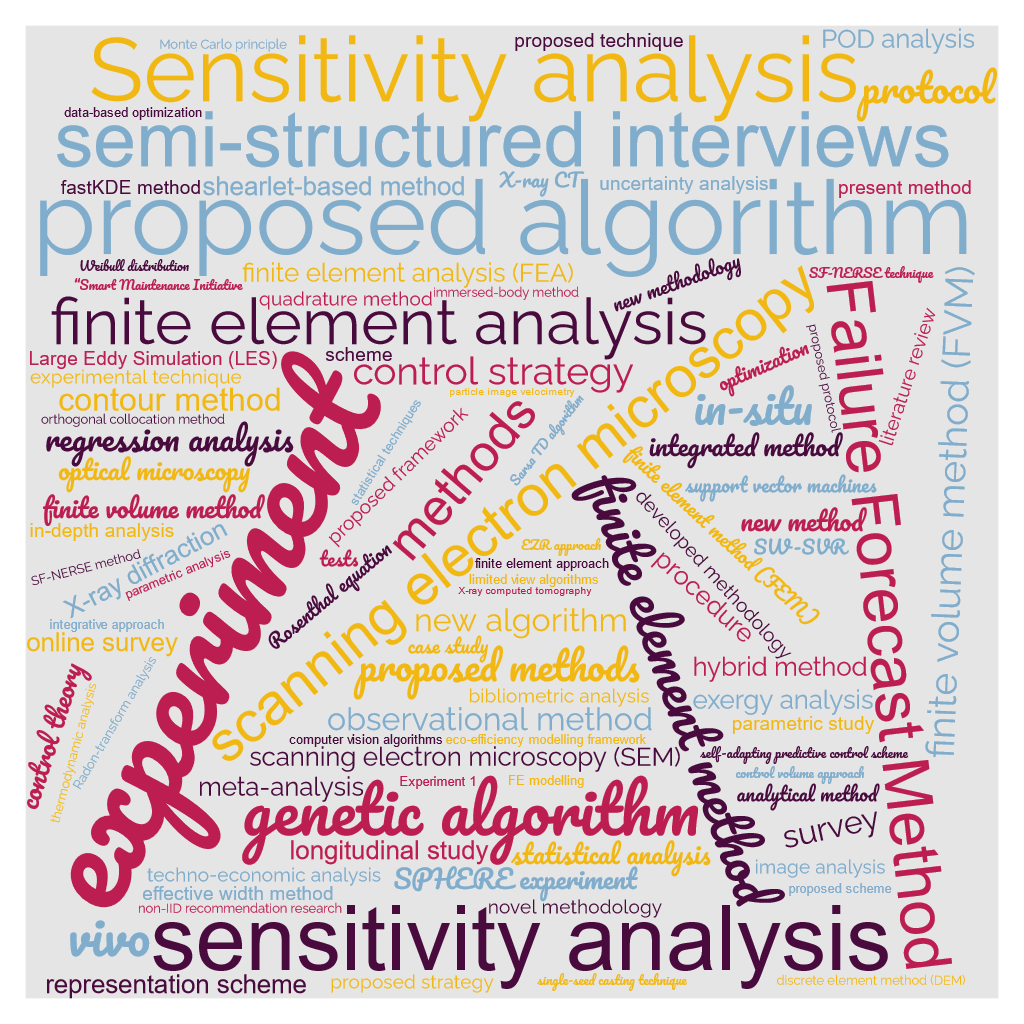} }}
    \subfloat[\textsc{material}]{{\includegraphics[width=3.9cm]{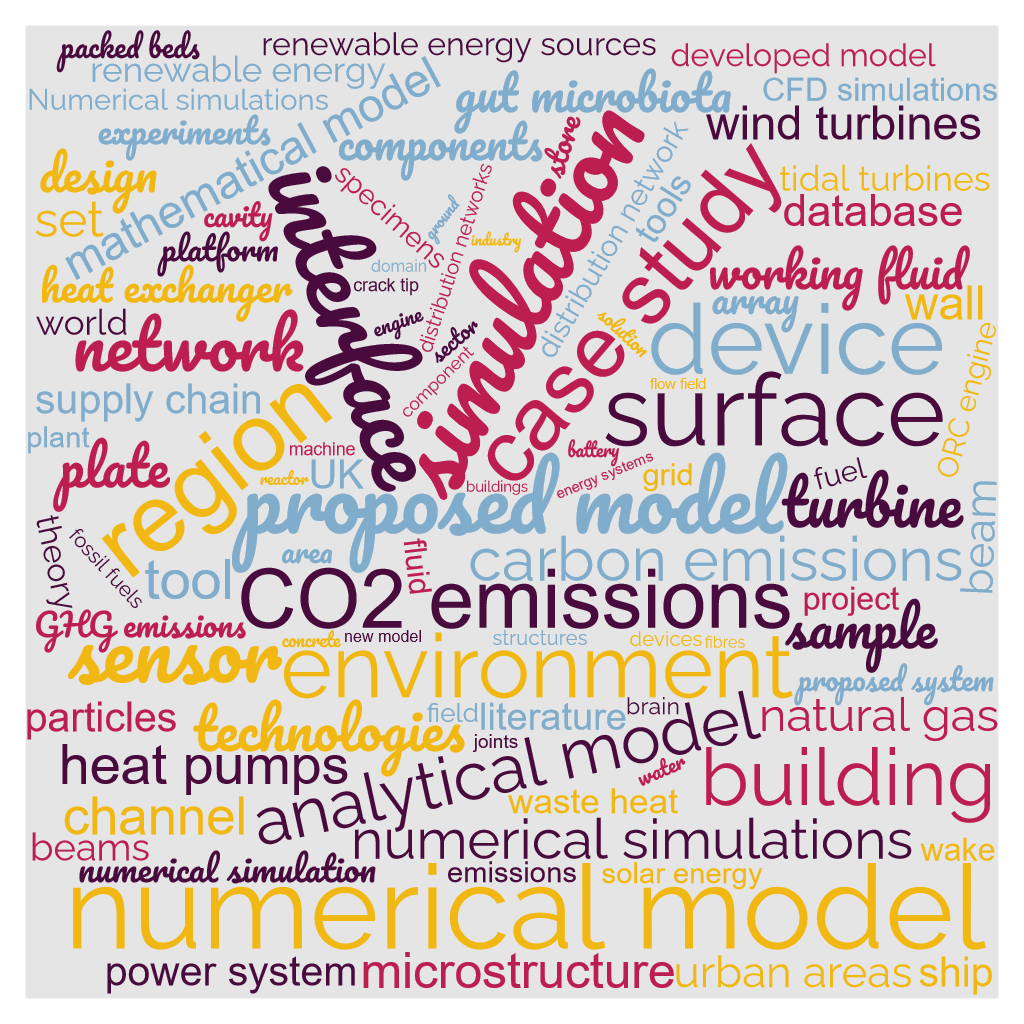} }}
    \subfloat[\textsc{data}]{{\includegraphics[width=3.9cm]{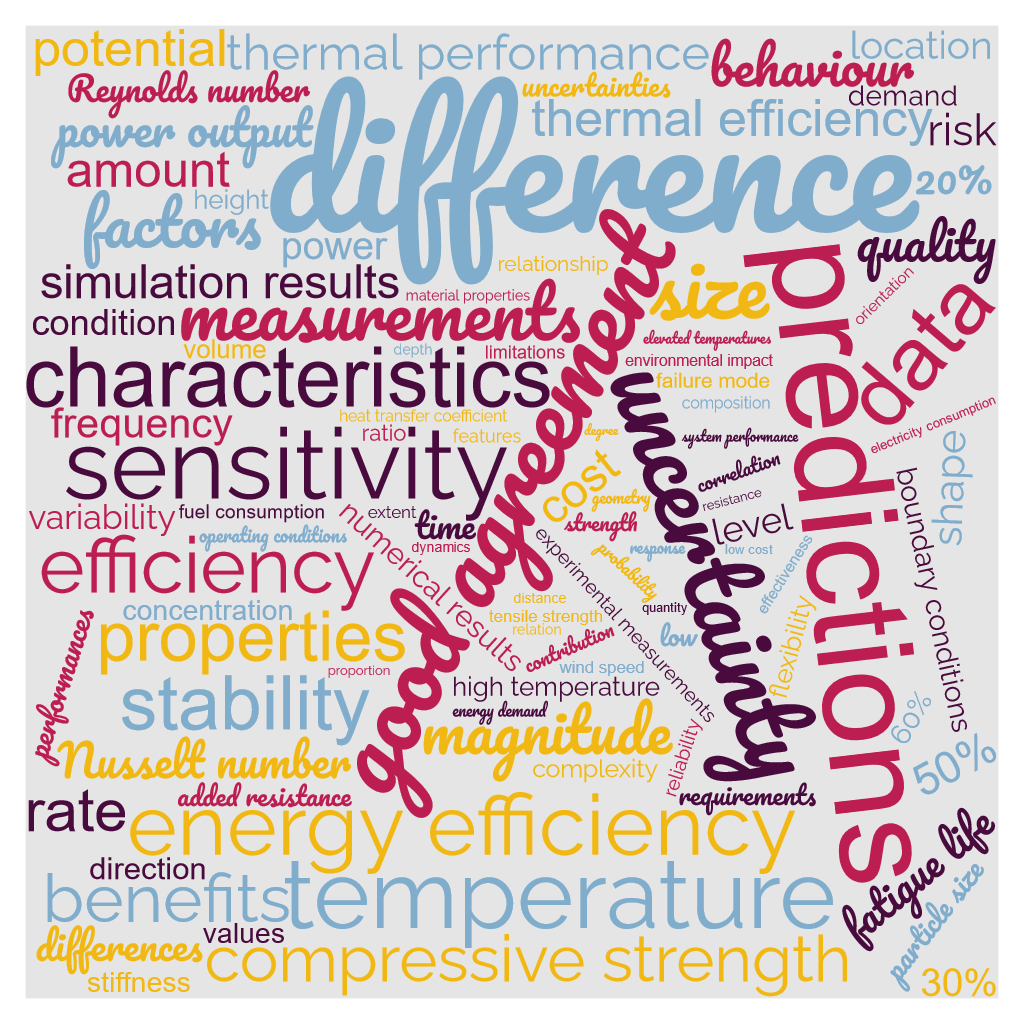} }}    
    \caption{Engineering domain word clouds}
    \label{fig:eng}
\end{figure*}

\begin{figure*}[!htb]
    \centering
    \subfloat[\textsc{process}]{{\includegraphics[width=3.9cm]{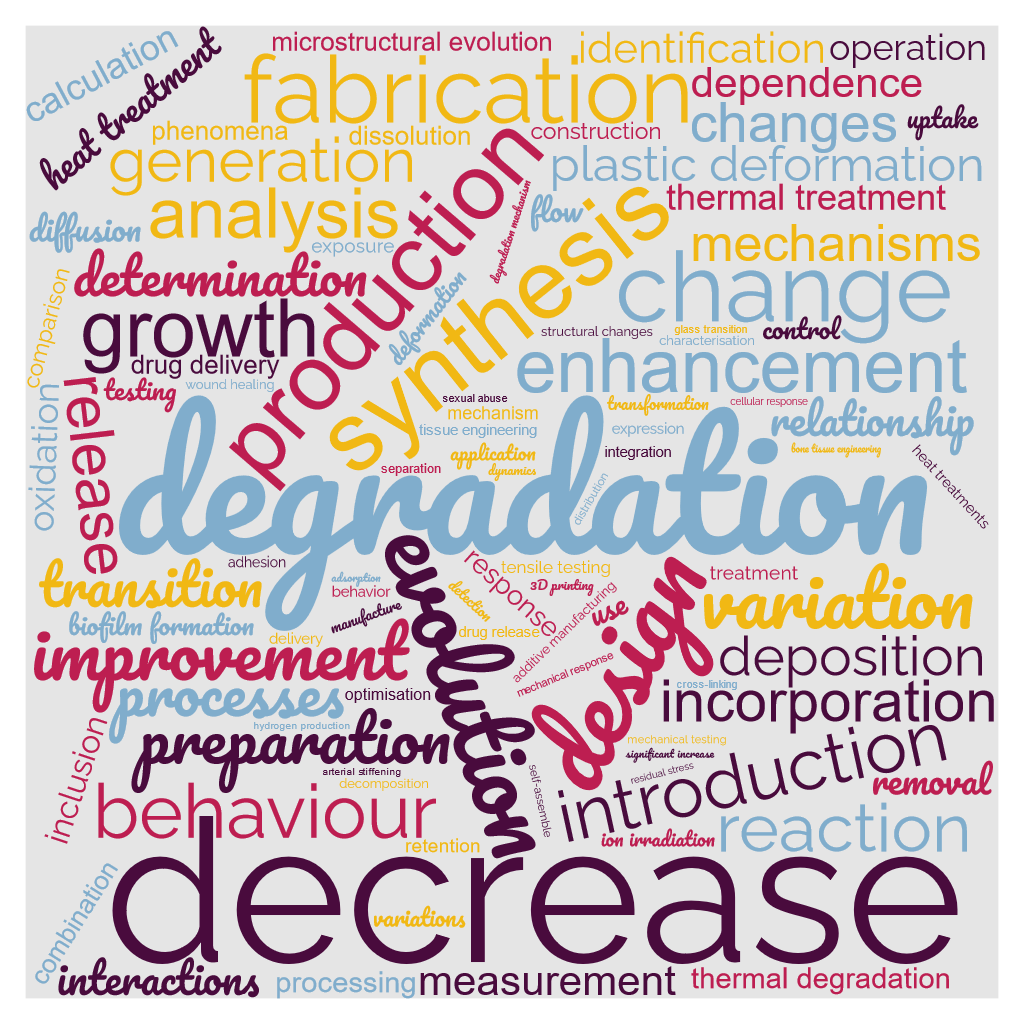} }}
    \subfloat[\textsc{method}]{{\includegraphics[width=3.9cm]{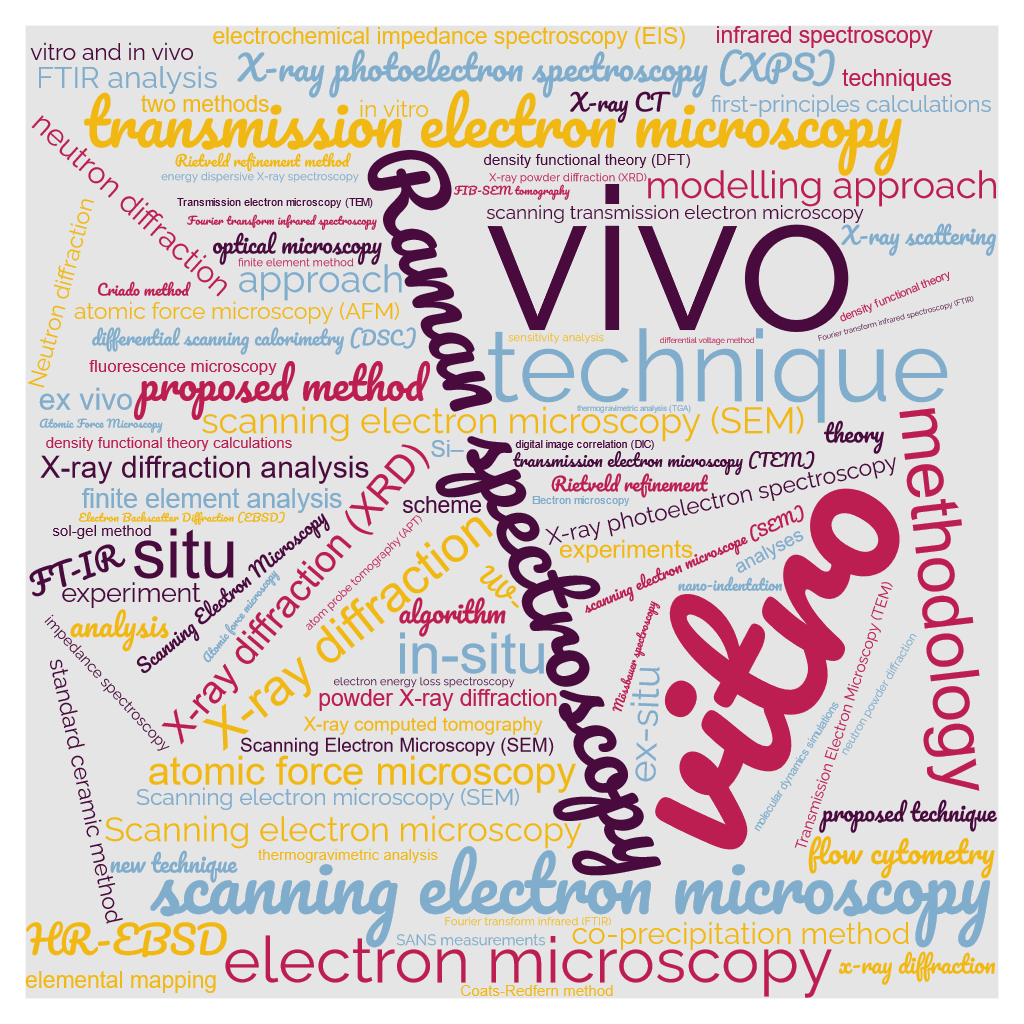} }}
    \subfloat[\textsc{material}]{{\includegraphics[width=3.9cm]{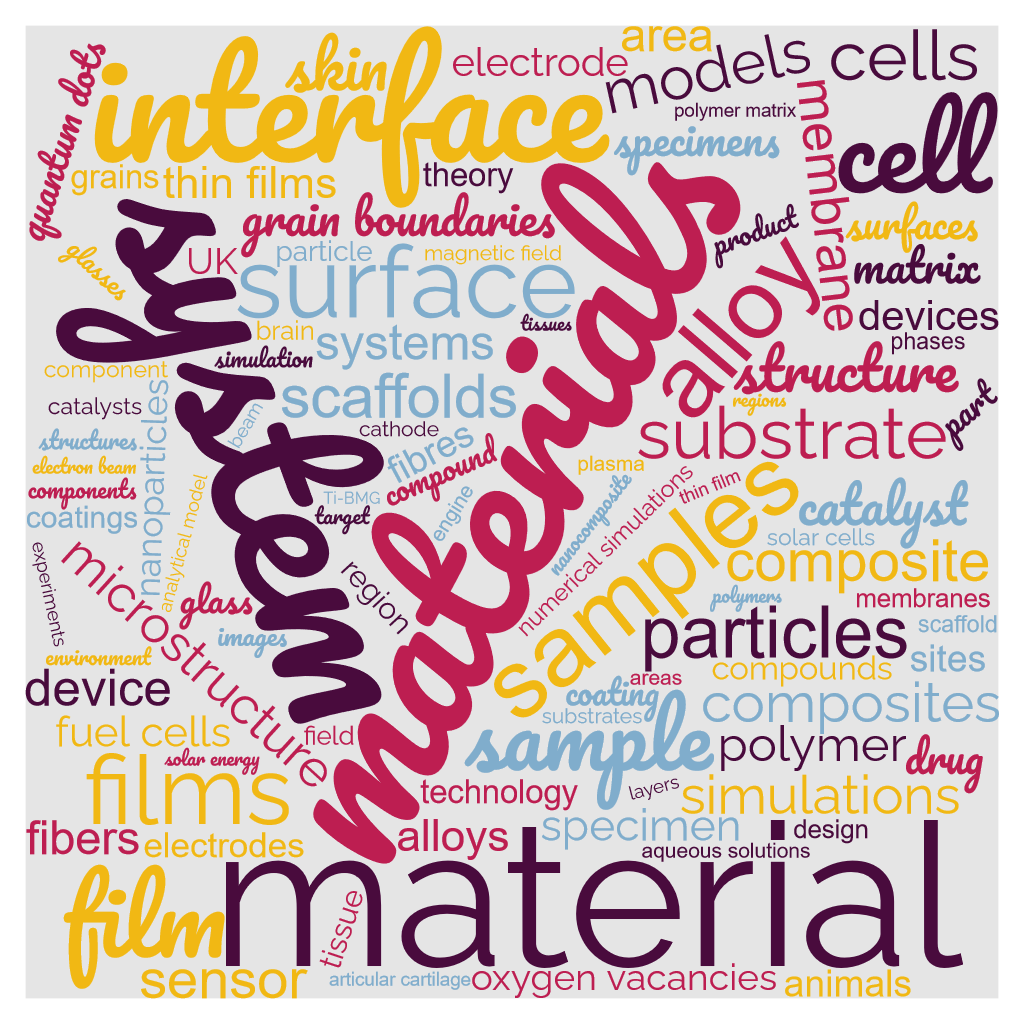} }}
    \subfloat[\textsc{data}]{{\includegraphics[width=3.9cm]{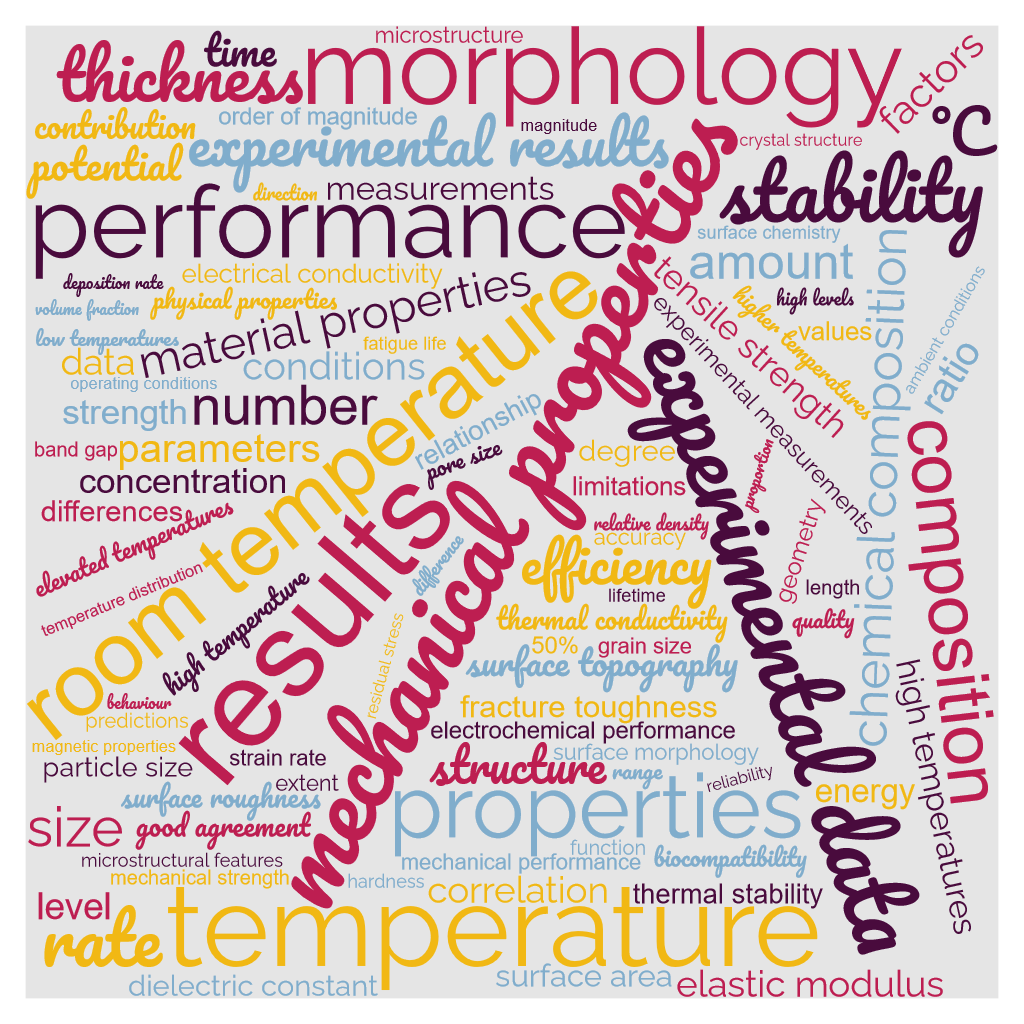} }}    
    \caption{Material Science domain word clouds}
    \label{fig:ms}
\end{figure*}

\begin{figure*}[!htb]
    \centering
    \subfloat[\textsc{process}]{{\includegraphics[width=3.9cm]{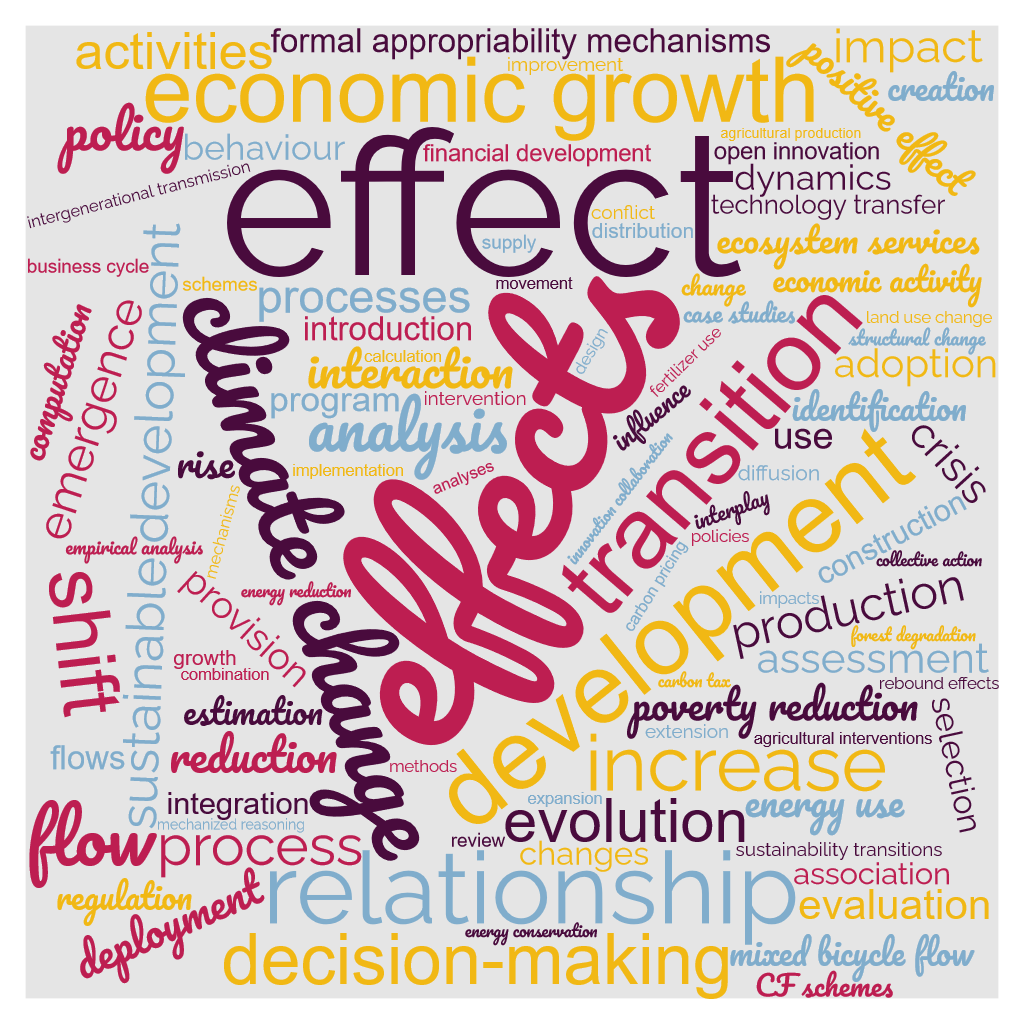} }}
    \subfloat[\textsc{method}]{{\includegraphics[width=3.9cm]{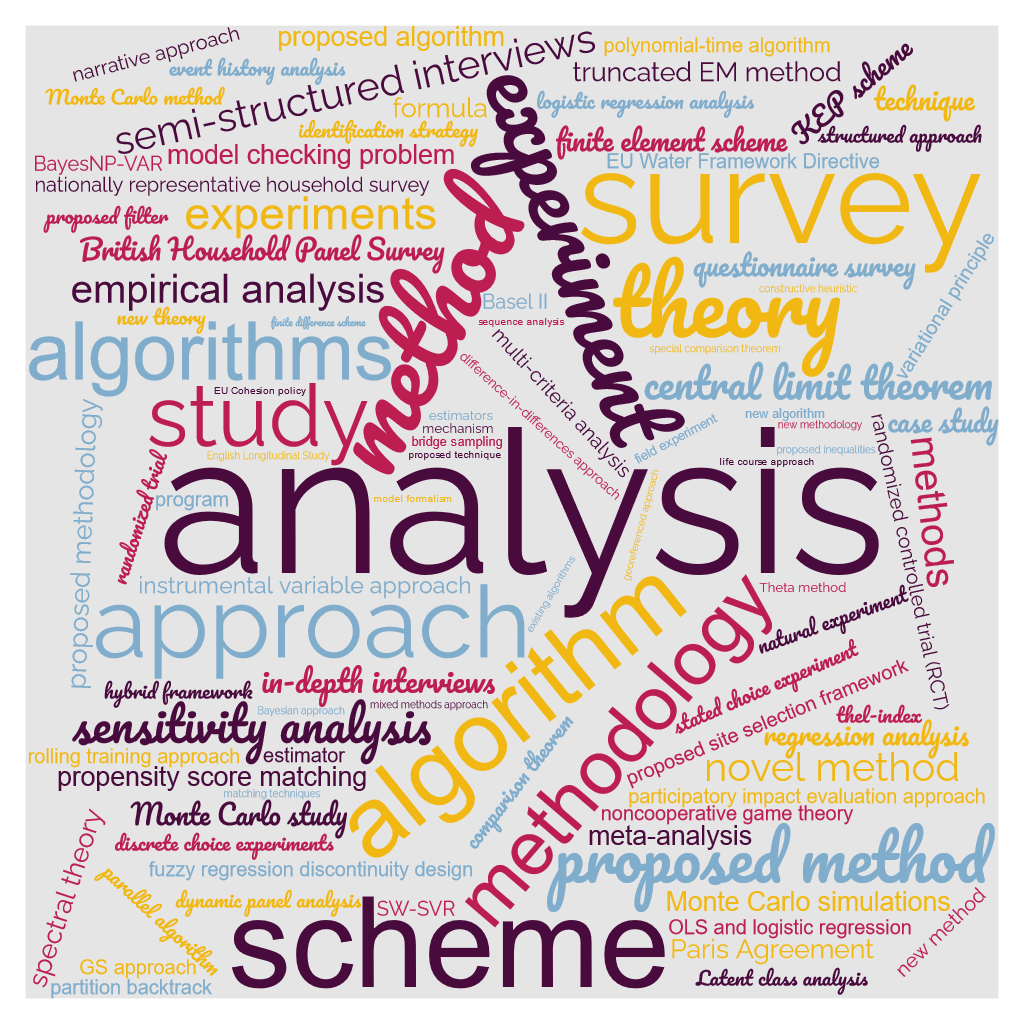} }}
    \subfloat[\textsc{material}]{{\includegraphics[width=3.9cm]{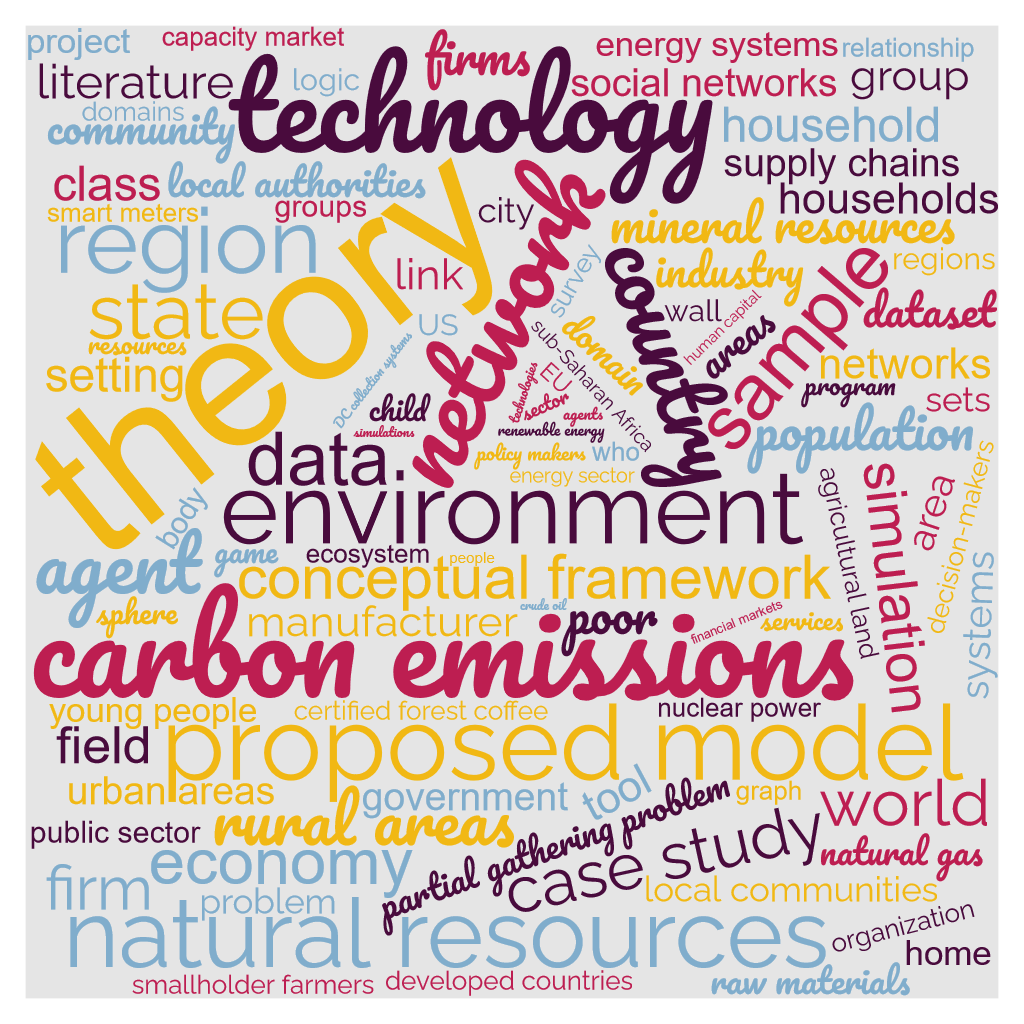} }}
    \subfloat[\textsc{data}]{{\includegraphics[width=3.9cm]{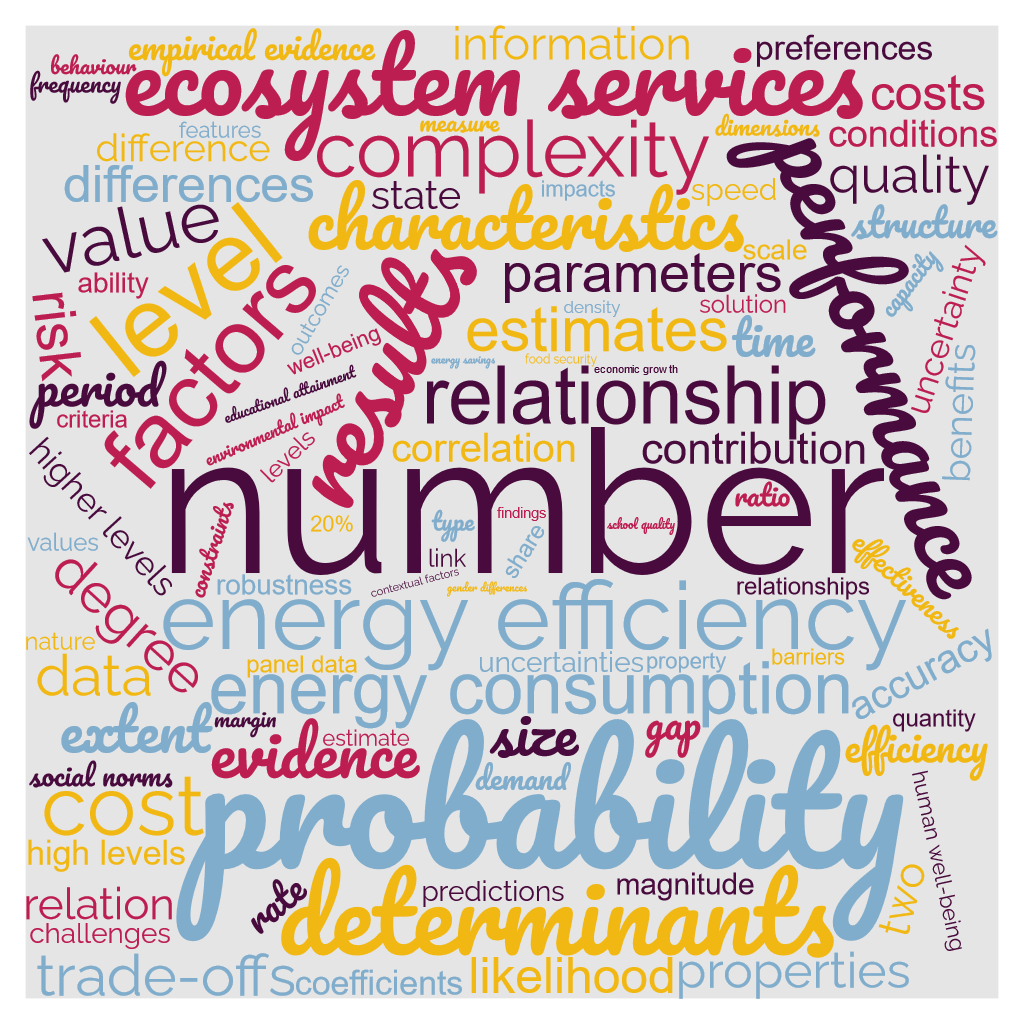} }}    
    \caption{Mathematics domain word clouds}
    \label{fig:math}
\end{figure*}

\begin{figure*}[!htb]
    \centering
    \subfloat[\textsc{process}]{{\includegraphics[width=3.9cm]{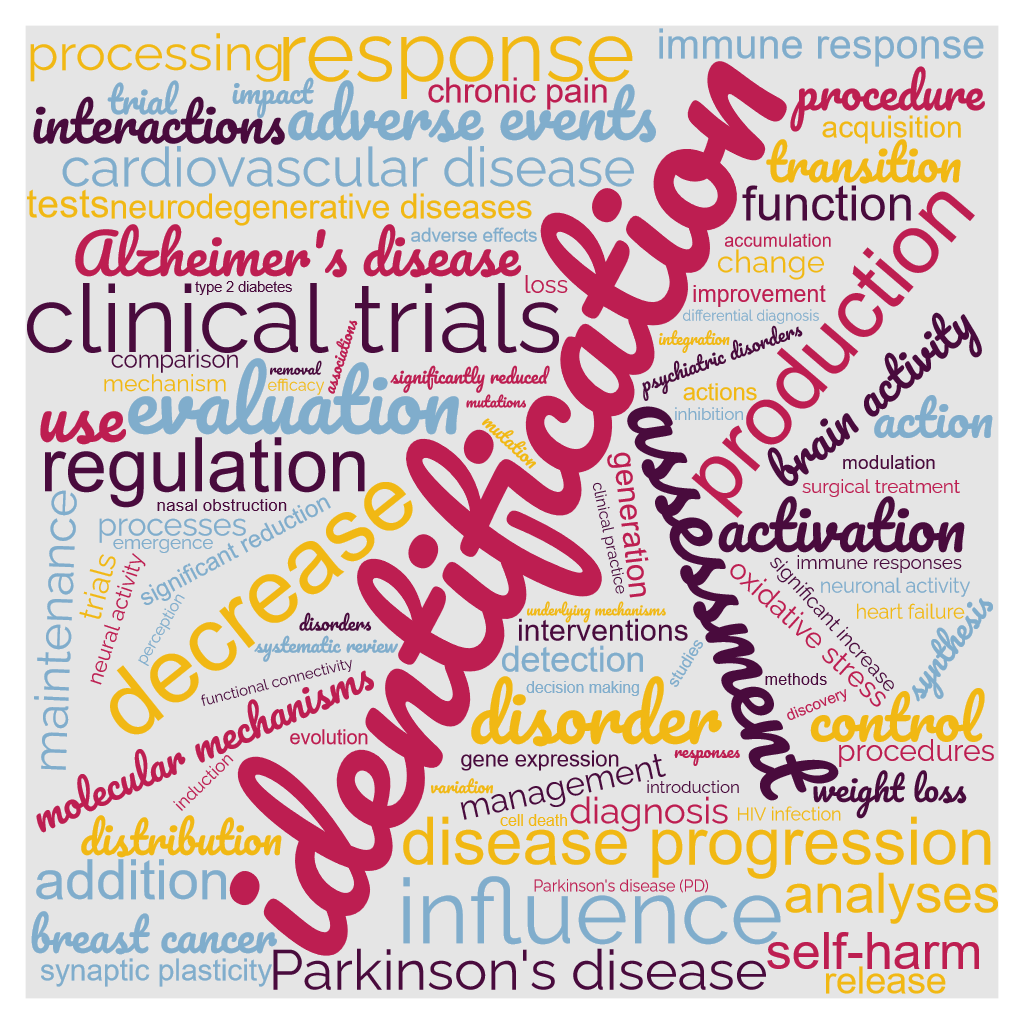} }}
    \subfloat[\textsc{method}]{{\includegraphics[width=3.9cm]{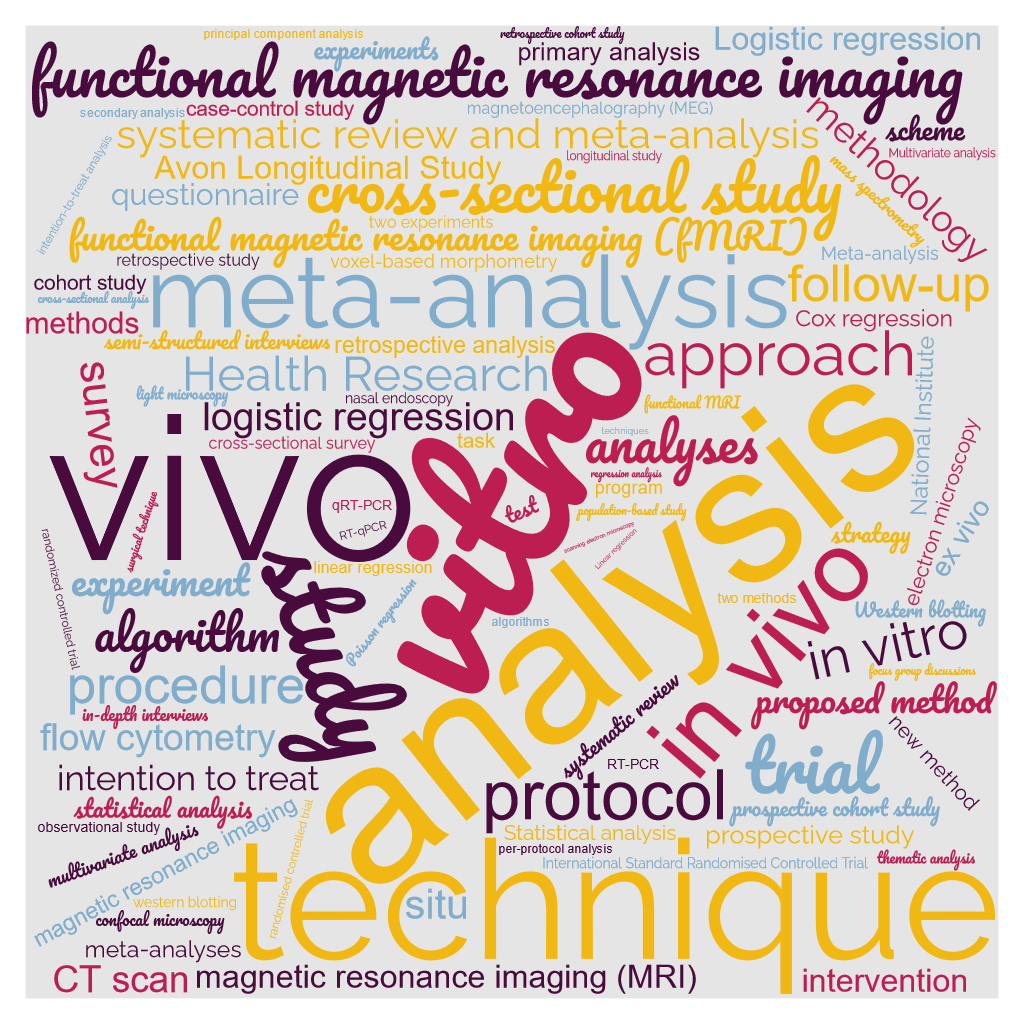} }}
    \subfloat[\textsc{material}]{{\includegraphics[width=3.9cm]{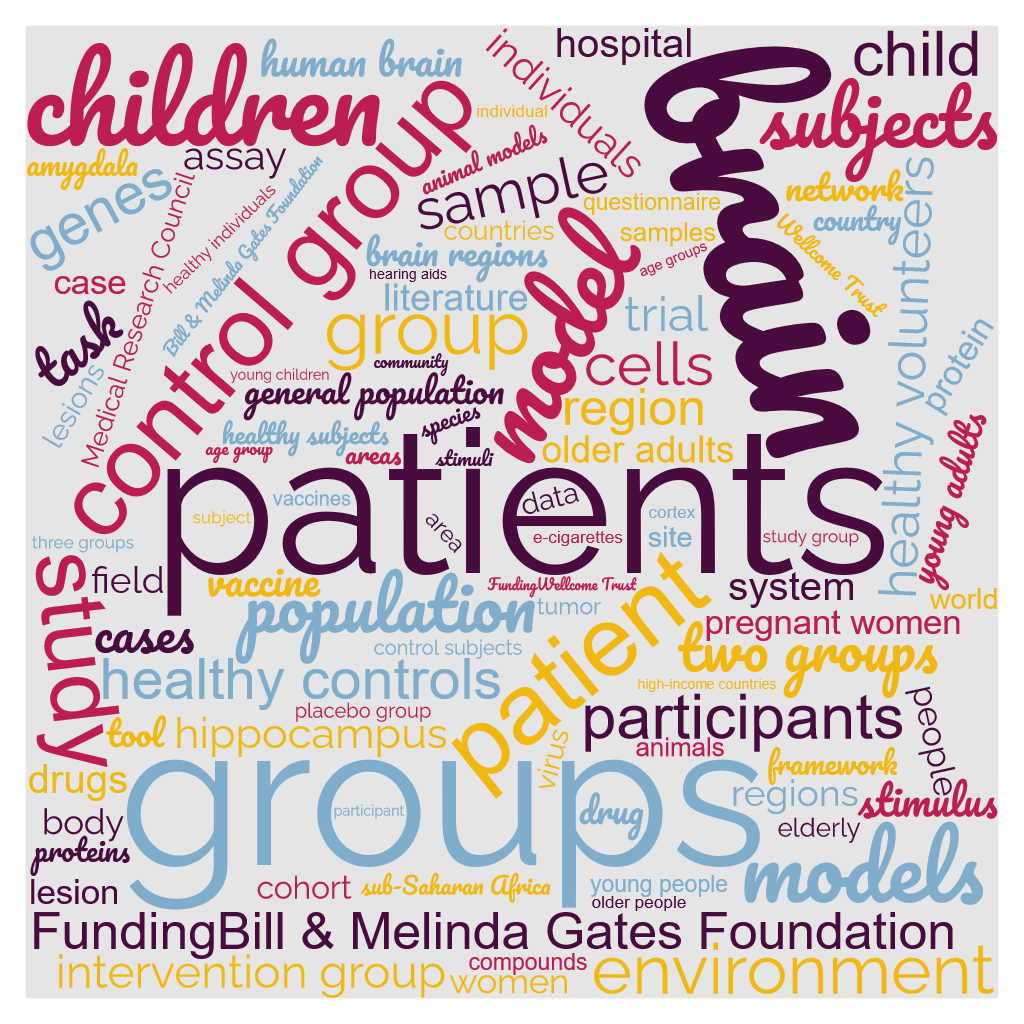} }}
    \subfloat[\textsc{data}]{{\includegraphics[width=3.9cm]{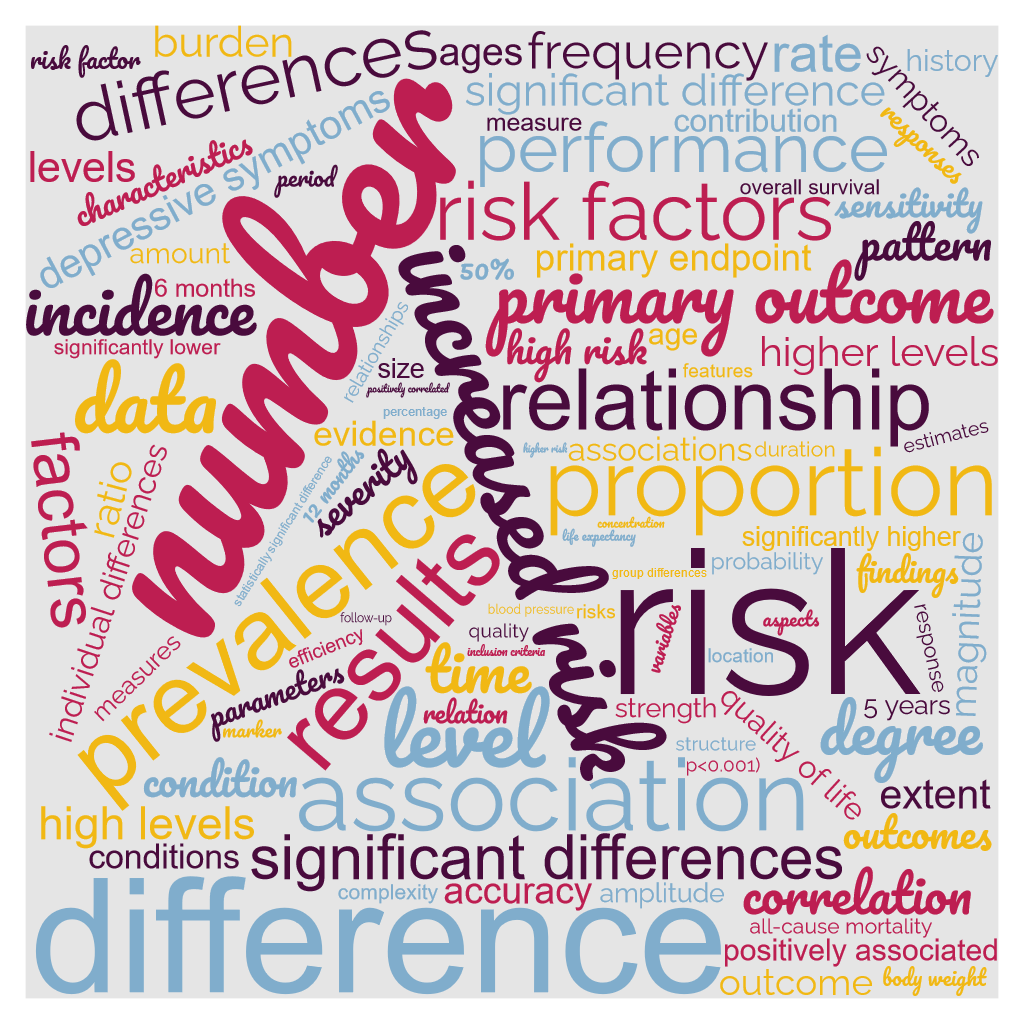} }}    
    \caption{Medicine domain word clouds}
    \label{fig:med}
\end{figure*}

\section{Applications}
\label{sec:app}

In this section, we offer some directions for practical applications developed based on STEM-ECR NER. With this we hope to offer inspiration for similar or related usages of the STEM-NER-60k corpus developed in this work.

\subsection{STEM Entities Recommendation Service in the Open Research Knowledge Graph}

\begin{figure*}[!htb]
\centering
\includegraphics[width=1.1\linewidth]{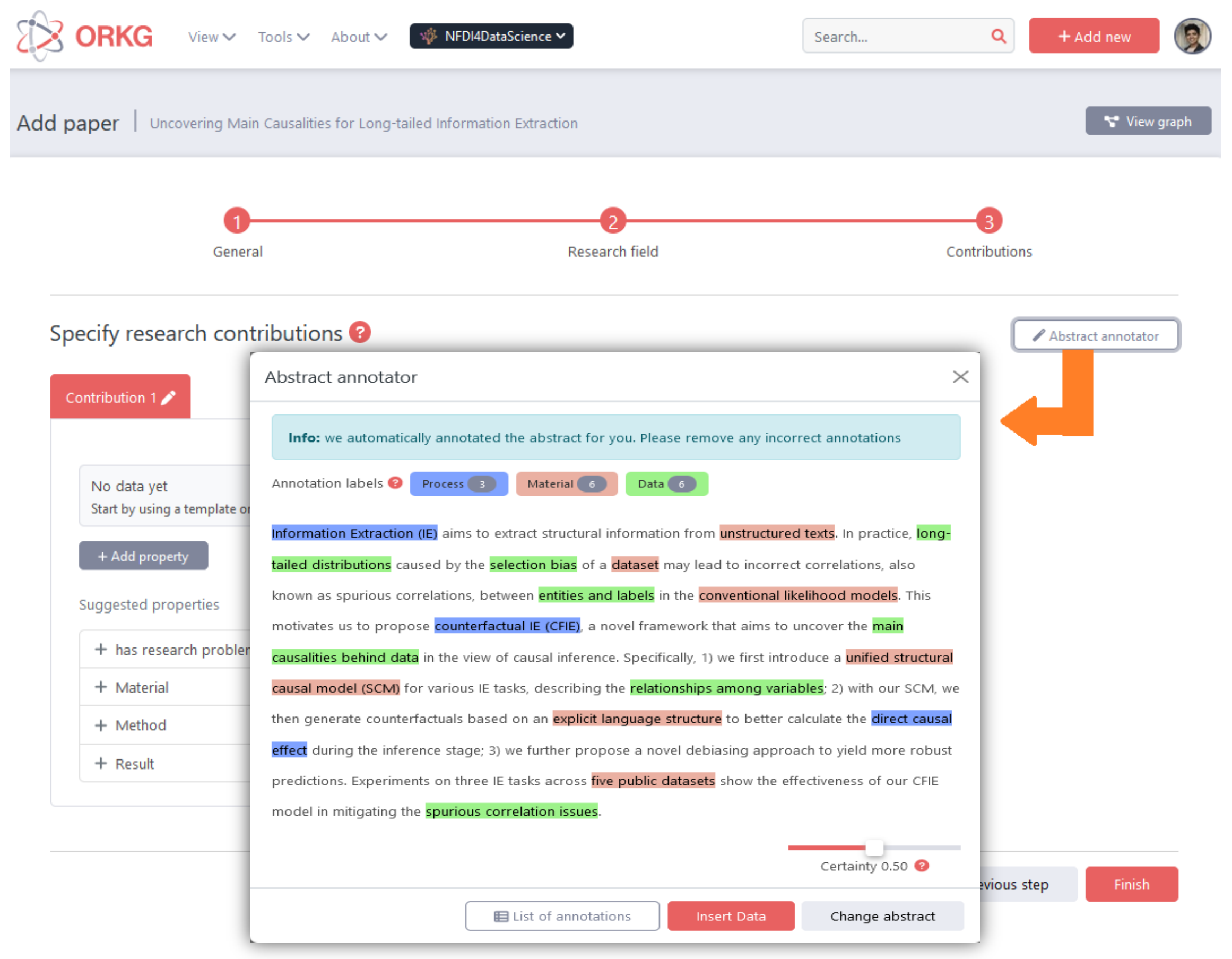}
\caption{STEM Entities-based Abstract Annotator Recommendation Service for \textsc{process}, \textsc{method}, \textsc{material}, and \textsc{data} entities in the Add-paper wizard of the next-generation Open Research Knowledge Graph (ORKG) digital library front-end service.}
\label{fig:app1}
\end{figure*}

The STEM-ECR annotation project (\url{https://data.uni-hannover.de/dataset/stem-ecr-v1-0}) was initiated to support the population of structured scientific concepts in the Open Research Knowledge Graph (ORKG). \autoref{fig:app1} demonstrates the integration of our prior developed machine learning model of STEM entities~\cite{brack2020domain} in the ORKG frontend. The model takes as input the Abstract of a new incoming publication and structures the Abstract per the four concepts. The user can then flexibly select and deselect annotations based on their prediction confidences or the user preference to automatically structure the contribution data of the work.


\subsection{Knowledge Graph Construction for Fine-Grained Structured Search}
\label{sec:kg}

\begin{figure}[!htb]
\centering
\includegraphics[width=\linewidth]{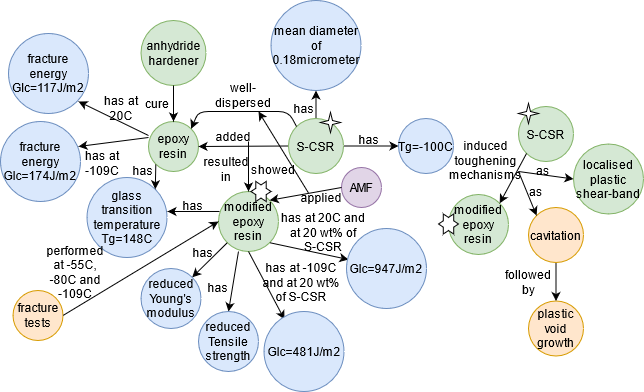}
\caption{Structured Knowledge Graph (KG) representation of a \textit{Material Science} domain publication Abstract~\cite{chen2013mechanical} as \textsc{process}, \textsc{method}, \textsc{material}, and \textsc{data} typed entities. For KGs in the remaining 9 STEM domains we consider, see \autoref{sec:appendix}.}
\label{fig:kg}
\end{figure}

Knowledge Graphs (KG) play a crucial role in many modern applications (\url{https://developers.google.com/knowledge-graph}) as solutions to the information access and search problem. There have been several initiatives in the NLP~\cite{semanticscholar,sciie,jdis,ncg}, and the Semantic Web~\cite{auer2018towards,orkg-old}, communities suggesting an increasing trend toward adoption of KGs for scientific articles. The automatic construction of KGs from text is a challenging problem, more so owing to the multidisciplinary nature of Science at large. While machines can better handle the volume of scientific literature, they need supervisory signals to determine which elements of the text have value. The STEM-NER-60k corpus can be leveraged to construct knowledge graphs underlying fine-grained search over publications. \autoref{fig:kg} demonstrates an example KG that was manually annotated with relations between the entities. In the figure, the entity nodes are color-coded by their concept type: orange corresponds to \textsc{process}, purple for \textsc{method}, green corresponds to \textsc{material}, and blue for \textsc{data}.


\section{Conclusion and Future Work Directions}

In this paper, we have systematically introduced a large-scale multidisciplinary STEM corpus spanning 10 disciplines with structured abstracts in terms of generic \textsc{process}, \textsc{method}, \textsc{material}, and \textsc{data} entities. The corpus called STEM-NER-60k is publicly released for facilitating future research \url{https://github.com/jd-coderepos/stem-ner-60k}. Based on the applications presented in \autoref{sec:app}, we envision two main future research directions: 1) extending the set of generic concepts beyond the four proposed in this work. E.g., our prior work~\cite{brack2020domain} proposed concepts such as \textsc{task}, \textsc{objective}, and \textsc{result}. And, 2) for automatic KG construction multidisciplinarily, annotating semantic relations between the entities.


\bibstyle{acl_natbib}
\bibliography{acl}

\begin{thebibliography}{76}
\expandafter\ifx\csname natexlab\endcsname\relax\def\natexlab#1{#1}\fi

\bibitem[{Adel(2018)}]{adel2018deep}
Heike Adel. 2018.
\newblock \emph{Deep learning methods for knowledge base population}.
\newblock Ph.D. thesis, lmu.

\bibitem[{Ammar et~al.(2018)Ammar, Groeneveld, Bhagavatula, Beltagy, Crawford,
  Downey, Dunkelberger, Elgohary, Feldman, Ha et~al.}]{semanticscholar}
Waleed Ammar, Dirk Groeneveld, Chandra Bhagavatula, Iz~Beltagy, Miles Crawford,
  Doug Downey, Jason Dunkelberger, Ahmed Elgohary, Sergey Feldman, Vu~Ha,
  et~al. 2018.
\newblock Construction of the literature graph in semantic scholar.
\newblock In \emph{NAACL-HLT (3)}.

\bibitem[{Ashburner et~al.(2000)Ashburner, Ball, Blake, Botstein, Butler,
  Cherry, Davis, Dolinski, Dwight, Eppig et~al.}]{geneontology}
Michael Ashburner, Catherine~A Ball, Judith~A Blake, David Botstein, Heather
  Butler, J~Michael Cherry, Allan~P Davis, Kara Dolinski, Selina~S Dwight,
  Janan~T Eppig, et~al. 2000.
\newblock Gene ontology: tool for the unification of biology.
\newblock \emph{Nature genetics}, 25(1):25--29.

\bibitem[{Auer et~al.(2018)Auer, Kovtun, Prinz, Kasprzik, Stocker, and
  Vidal}]{auer2018towards}
S{\"o}ren Auer, Viktor Kovtun, Manuel Prinz, Anna Kasprzik, Markus Stocker, and
  Maria~Esther Vidal. 2018.
\newblock Towards a knowledge graph for science.
\newblock In \emph{Proceedings of the 8th International Conference on Web
  Intelligence, Mining and Semantics}, pages 1--6.

\bibitem[{Auer et~al.(2020)Auer, Oelen, Haris, Stocker, D’Souza, Farfar,
  Vogt, Prinz, Wiens, and Jaradeh}]{auer2020improving}
S{\"o}ren Auer, Allard Oelen, Muhammad Haris, Markus Stocker, Jennifer
  D’Souza, Kheir~Eddine Farfar, Lars Vogt, Manuel Prinz, Vitalis Wiens, and
  Mohamad~Yaser Jaradeh. 2020.
\newblock Improving access to scientific literature with knowledge graphs.
\newblock \emph{Bibliothek Forschung und Praxis}, 44(3):516--529.

\bibitem[{Auer(2018{\natexlab{a}})}]{auer_soren_2018}
Sören Auer. 2018{\natexlab{a}}.
\newblock \href {https://doi.org/10.5281/zenodo.1157185} {Towards an open
  research knowledge graph}.

\bibitem[{Auer(2018{\natexlab{b}})}]{auer_orkg}
Sören Auer. 2018{\natexlab{b}}.
\newblock \href {https://doi.org/10.5281/zenodo.1157185} {Towards an open
  research knowledge graph}.

\bibitem[{Augenstein et~al.(2017{\natexlab{a}})Augenstein, Das, Riedel,
  Vikraman, and McCallum}]{augenstein2017semeval}
Isabelle Augenstein, Mrinal Das, Sebastian Riedel, Lakshmi Vikraman, and Andrew
  McCallum. 2017{\natexlab{a}}.
\newblock Semeval 2017 task 10: Scienceie - extracting keyphrases and relations
  from scientific publications.
\newblock In \emph{SemEval@ACL}.

\bibitem[{Augenstein et~al.(2017{\natexlab{b}})Augenstein, Das, Riedel,
  Vikraman, and McCallum}]{scienceie}
Isabelle Augenstein, Mrinal Das, Sebastian Riedel, Lakshmi Vikraman, and Andrew
  McCallum. 2017{\natexlab{b}}.
\newblock \href {https://doi.org/10.18653/v1/S17-2091} {{S}em{E}val 2017 task
  10: {S}cience{IE} - extracting keyphrases and relations from scientific
  publications}.
\newblock In \emph{Proceedings of the 11th International Workshop on Semantic
  Evaluation ({S}em{E}val-2017)}, pages 546--555, Vancouver, Canada.
  Association for Computational Linguistics.

\bibitem[{Bada et~al.(2012)Bada, Eckert, Evans, Garcia, Shipley, Sitnikov,
  Baumgartner, Cohen, Verspoor, Blake et~al.}]{craft}
Michael Bada, Miriam Eckert, Donald Evans, Kristin Garcia, Krista Shipley,
  Dmitry Sitnikov, William~A Baumgartner, K~Bretonnel Cohen, Karin Verspoor,
  Judith~A Blake, et~al. 2012.
\newblock Concept annotation in the craft corpus.
\newblock \emph{BMC bioinformatics}, 13(1):1--20.

\bibitem[{Bader et~al.(2006)Bader, Cary, and Sander}]{pathguide}
Gary~D Bader, Michael~P Cary, and Chris Sander. 2006.
\newblock Pathguide: a pathway resource list.
\newblock \emph{Nucleic acids research}, 34(suppl\_1):D504--D506.

\bibitem[{Beltagy et~al.(2019)Beltagy, Lo, and Cohan}]{beltagy2019scibert}
Iz~Beltagy, Kyle Lo, and Arman Cohan. 2019.
\newblock Scibert: A pretrained language model for scientific text.
\newblock In \emph{EMNLP-IJCNLP}, pages 3606--3611.

\bibitem[{Birkle et~al.(2020)Birkle, Pendlebury, Schnell, and
  Adams}]{birkle2020web}
Caroline Birkle, David~A Pendlebury, Joshua Schnell, and Jonathan Adams. 2020.
\newblock Web of science as a data source for research on scientific and
  scholarly activity.
\newblock \emph{Quantitative Science Studies}, 1(1):363--376.

\bibitem[{Bodenreider(2004)}]{umls}
Olivier Bodenreider. 2004.
\newblock The unified medical language system (umls): integrating biomedical
  terminology.
\newblock \emph{Nucleic Acids Research}, 32(Database issue):D267.

\bibitem[{Brack et~al.(2020)Brack, D’Souza, Hoppe, Auer, and
  Ewerth}]{brack2020domain}
Arthur Brack, Jennifer D’Souza, Anett Hoppe, S{\"o}ren Auer, and Ralph
  Ewerth. 2020.
\newblock Domain-independent extraction of scientific concepts from research
  articles.
\newblock In \emph{European Conference on Information Retrieval}, pages
  251--266. Springer.

\bibitem[{Camon et~al.(2004)Camon, Magrane, Barrell, Lee, Dimmer, Maslen,
  Binns, Harte, Lopez, and Apweiler}]{goa}
Evelyn Camon, Michele Magrane, Daniel Barrell, Vivian Lee, Emily Dimmer, John
  Maslen, David Binns, Nicola Harte, Rodrigo Lopez, and Rolf Apweiler. 2004.
\newblock The gene ontology annotation (goa) database: sharing knowledge in
  uniprot with gene ontology.
\newblock \emph{Nucleic Acids Research}, 32(Database issue):D262.

\bibitem[{Chatr-Aryamontri et~al.(2007)Chatr-Aryamontri, Ceol, Palazzi,
  Nardelli, Schneider, Castagnoli, and Cesareni}]{mint}
Andrew Chatr-Aryamontri, Arnaud Ceol, Luisa~Montecchi Palazzi, Giuliano
  Nardelli, Maria~Victoria Schneider, Luisa Castagnoli, and Gianni Cesareni.
  2007.
\newblock Mint: the molecular interaction database.
\newblock \emph{Nucleic acids research}, 35(suppl\_1):D572--D574.

\bibitem[{Chen et~al.(2013)Chen, Kinloch, Sprenger, and
  Taylor}]{chen2013mechanical}
J~Chen, AJ~Kinloch, S~Sprenger, and AC~Taylor. 2013.
\newblock The mechanical properties and toughening mechanisms of an epoxy
  polymer modified with polysiloxane-based core-shell particles.
\newblock \emph{Polymer}, 54(16):4276--4289.

\bibitem[{Collier and Kim(2004)}]{bionlp}
Nigel Collier and Jin-Dong Kim. 2004.
\newblock \href {https://aclanthology.org/W04-1213} {Introduction to the
  bio-entity recognition task at {JNLPBA}}.
\newblock In \emph{Proceedings of the International Joint Workshop on Natural
  Language Processing in Biomedicine and its Applications
  ({NLPBA}/{B}io{NLP})}, pages 73--78, Geneva, Switzerland. COLING.

\bibitem[{Corbett et~al.(2007)Corbett, Batchelor, and Teufel}]{chemnerdata}
Peter Corbett, Colin Batchelor, and Simone Teufel. 2007.
\newblock Annotation of chemical named entities.
\newblock In \emph{Biological, translational, and clinical language
  processing}, pages 57--64.

\bibitem[{Cotton et~al.(2014)Cotton, Patchigolla, and Oakey}]{chem-kg}
Alissa Cotton, Kumar Patchigolla, and John~E Oakey. 2014.
\newblock Minor and trace element emissions from post-combustion co2 capture
  from coal: Experimental and equilibrium calculations.
\newblock \emph{Fuel}, 117:391--407.

\bibitem[{Cover(1999)}]{cover}
Thomas~M Cover. 1999.
\newblock \emph{Elements of information theory}.
\newblock John Wiley \& Sons.

\bibitem[{Devlin et~al.(2018)Devlin, Chang, Lee, and Toutanova}]{devlin}
Jacob Devlin, Ming-Wei Chang, Kenton Lee, and Kristina Toutanova. 2018.
\newblock Bert: Pre-training of deep bidirectional transformers for language
  understanding.
\newblock \emph{CoRR}, abs/1810.04805.

\bibitem[{Do{\u{g}}an et~al.(2014)Do{\u{g}}an, Leaman, and
  Lu}]{ncbidiseasecorpus}
Rezarta~Islamaj Do{\u{g}}an, Robert Leaman, and Zhiyong Lu. 2014.
\newblock Ncbi disease corpus: a resource for disease name recognition and
  concept normalization.
\newblock \emph{Journal of biomedical informatics}, 47:1--10.

\bibitem[{Dolev et~al.(2014)Dolev, F{\"u}gger, Posch, Schmid, Steininger, and
  Lenzen}]{cs-kg}
Danny Dolev, Matthias F{\"u}gger, Markus Posch, Ulrich Schmid, Andreas
  Steininger, and Christoph Lenzen. 2014.
\newblock Rigorously modeling self-stabilizing fault-tolerant circuits: An
  ultra-robust clocking scheme for systems-on-chip.
\newblock \emph{Journal of Computer and System Sciences}, 80(4):860--900.

\bibitem[{D'Souza and Auer(2021)}]{cl-titles}
Jennifer D'Souza and Soeren Auer. 2021.
\newblock Pattern-based acquisition of scientific entities from scholarly
  article titles.
\newblock \emph{arXiv preprint arXiv:2109.00199}.

\bibitem[{D{'}Souza et~al.(2020)D{'}Souza, Hoppe, Brack, Jaradeh, Auer, and
  Ewerth}]{stem-ecr}
Jennifer D{'}Souza, Anett Hoppe, Arthur Brack, Mohmad~Yaser Jaradeh, S{\"o}ren
  Auer, and Ralph Ewerth. 2020.
\newblock \href {https://aclanthology.org/2020.lrec-1.268} {The {STEM}-{ECR}
  dataset: Grounding scientific entity references in {STEM} scholarly content
  to authoritative encyclopedic and lexicographic sources}.
\newblock In \emph{Proceedings of the 12th Language Resources and Evaluation
  Conference}, pages 2192--2203, Marseille, France. European Language Resources
  Association.

\bibitem[{D’Souza and Auer(2021)}]{jdis}
Jennifer D’Souza and S{\"o}ren Auer. 2021.
\newblock Sentence, phrase, and triple annotations to build a knowledge graph
  of natural language processing contributions—a trial dataset.
\newblock \emph{Journal of Data and Information Science}, 6(3):6--34.

\bibitem[{D’Souza et~al.(2021)D’Souza, Auer, and Pedersen}]{ncg}
Jennifer D’Souza, S{\"o}ren Auer, and Ted Pedersen. 2021.
\newblock Semeval-2021 task 11: Nlpcontributiongraph-structuring scholarly nlp
  contributions for a research knowledge graph.
\newblock In \emph{Proceedings of the 15th International Workshop on Semantic
  Evaluation (SemEval-2021)}, pages 364--376.

\bibitem[{D’Souza et~al.(2020)D’Souza, Hoppe, Brack, Jaradeh, Auer, and
  Ewerth}]{dsouza2020stemecr}
Jennifer D’Souza, Anett Hoppe, Arthur Brack, Mohmad~Yaser Jaradeh, S{\"o}ren
  Auer, and Ralph Ewerth. 2020.
\newblock The stem-ecr dataset: Grounding scientific entity references in stem
  scholarly content to authoritative encyclopedic and lexicographic sources.
\newblock In \emph{Proceedings of The 12th Language Resources and Evaluation
  Conference}, pages 2192--2203.

\bibitem[{F{\"{a}}rber et~al.(2021)F{\"{a}}rber, Albers, and
  Schüber}]{used_meth_dataset}
Michael F{\"{a}}rber, Alexander Albers, and Felix Schüber. 2021.
\newblock Identifying used methods and datasets in scientific publications.
\newblock In \emph{Proceedings of the Workshop on Scientific Document
  Understanding: co-located with 35th AAAI Conference on Artificial Inteligence
  (AAAI 2021) ; Remote, February 9, 2021. Ed.: A. P. B. Veyseh}, volume 2831 of
  \emph{CEUR Workshop Proceedings}. {CEUR Workshop Proceedings (CEUR-WS)}.

\bibitem[{Fricke(2018)}]{fricke2018semantic}
Suzanne Fricke. 2018.
\newblock Semantic scholar.
\newblock \emph{Journal of the Medical Library Association: JMLA}, 106(1):145.

\bibitem[{G{\'a}bor et~al.(2018{\natexlab{a}})G{\'a}bor, Buscaldi, Schumann,
  QasemiZadeh, Zargayouna, and Charnois}]{gabor2018semeval}
Kata G{\'a}bor, Davide Buscaldi, Anne-Kathrin Schumann, Behrang QasemiZadeh,
  Haifa Zargayouna, and Thierry Charnois. 2018{\natexlab{a}}.
\newblock Semeval-2018 task 7: Semantic relation extraction and classification
  in scientific papers.
\newblock In \emph{SemEval}, pages 679--688.

\bibitem[{G{\'a}bor et~al.(2018{\natexlab{b}})G{\'a}bor, Buscaldi, Schumann,
  QasemiZadeh, Zargayouna, and Charnois}]{gabor-etal-2018-semeval}
Kata G{\'a}bor, Davide Buscaldi, Anne-Kathrin Schumann, Behrang QasemiZadeh,
  Ha{\"\i}fa Zargayouna, and Thierry Charnois. 2018{\natexlab{b}}.
\newblock \href {https://doi.org/10.18653/v1/S18-1111} {{S}em{E}val-2018 task
  7: Semantic relation extraction and classification in scientific papers}.
\newblock In \emph{Proceedings of The 12th International Workshop on Semantic
  Evaluation}, pages 679--688, New Orleans, Louisiana. Association for
  Computational Linguistics.

\bibitem[{Gangemi et~al.(2017)Gangemi, Presutti, Recupero, Nuzzolese,
  Draicchio, and MongiovÃ¬}]{fred}
Aldo Gangemi, Valentina Presutti, Diego~Reforgiato Recupero, Andrea~Giovanni
  Nuzzolese, Francesco Draicchio, and Misael MongiovÃ¬. 2017.
\newblock {Semantic Web Machine Reading with FRED}.
\newblock \emph{Semantic Web}, 8(6):873--893.

\bibitem[{Gupta and Manning(2011)}]{ftd}
Sonal Gupta and Christopher Manning. 2011.
\newblock \href {https://aclanthology.org/I11-1001} {Analyzing the dynamics of
  research by extracting key aspects of scientific papers}.
\newblock In \emph{Proceedings of 5th International Joint Conference on Natural
  Language Processing}, pages 1--9, Chiang Mai, Thailand. Asian Federation of
  Natural Language Processing.

\bibitem[{Haution(2012)}]{math-kg}
Olivier Haution. 2012.
\newblock Integrality of the chern character in small codimension.
\newblock \emph{Advances in Mathematics}, 231(2):855--878.

\bibitem[{Herrero-Zazo et~al.(2013)Herrero-Zazo, Segura-Bedmar, Mart{\'\i}nez,
  and Declerck}]{biocreativeDDI}
Mar{\'\i}a Herrero-Zazo, Isabel Segura-Bedmar, Paloma Mart{\'\i}nez, and
  Thierry Declerck. 2013.
\newblock The ddi corpus: An annotated corpus with pharmacological substances
  and drug--drug interactions.
\newblock \emph{Journal of biomedical informatics}, 46(5):914--920.

\bibitem[{Hou et~al.(2019)Hou, Jochim, Gleize, Bonin, and Ganguly}]{ibm-tdm}
Yufang Hou, Charles Jochim, Martin Gleize, Francesca Bonin, and Debasis
  Ganguly. 2019.
\newblock \href {https://doi.org/10.18653/v1/P19-1513} {Identification of
  tasks, datasets, evaluation metrics, and numeric scores for scientific
  leaderboards construction}.
\newblock In \emph{Proceedings of the 57th Annual Meeting of the Association
  for Computational Linguistics}, pages 5203--5213, Florence, Italy.
  Association for Computational Linguistics.

\bibitem[{Islamaj et~al.(2021)Islamaj, Leaman, Kim, Kwon, Wei, Comeau, Peng,
  Cissel, Coss, Fisher et~al.}]{nlmchem}
Rezarta Islamaj, Robert Leaman, Sun Kim, Dongseop Kwon, Chih-Hsuan Wei,
  Donald~C Comeau, Yifan Peng, David Cissel, Cathleen Coss, Carol Fisher,
  et~al. 2021.
\newblock Nlm-chem, a new resource for chemical entity recognition in pubmed
  full text literature.
\newblock \emph{Scientific Data}, 8(1):1--12.

\bibitem[{Jain et~al.(2020)Jain, van Zuylen, Hajishirzi, and Beltagy}]{scirex}
Sarthak Jain, Madeleine van Zuylen, Hannaneh Hajishirzi, and Iz~Beltagy. 2020.
\newblock \href {https://doi.org/10.18653/v1/2020.acl-main.670} {{S}ci{REX}:
  {A} challenge dataset for document-level information extraction}.
\newblock In \emph{Proceedings of the 58th Annual Meeting of the Association
  for Computational Linguistics}, pages 7506--7516, Online. Association for
  Computational Linguistics.

\bibitem[{Jaradeh et~al.(2019)Jaradeh, Oelen, Farfar, Prinz, D'Souza,
  Kismih\'{o}k, Stocker, and Auer}]{orkg-old}
Mohamad~Yaser Jaradeh, Allard Oelen, Kheir~Eddine Farfar, Manuel Prinz,
  Jennifer D'Souza, G\'{a}bor Kismih\'{o}k, Markus Stocker, and S\"{o}ren Auer.
  2019.
\newblock \href {https://doi.org/10.1145/3360901.3364435} {Open research
  knowledge graph: Next generation infrastructure for semantic scholarly
  knowledge}.
\newblock In \emph{Proceedings of the 10th International Conference on
  Knowledge Capture}, K-CAP '19, page 243–246, New York, NY, USA. Association
  for Computing Machinery.

\bibitem[{Kabongo et~al.(2021)Kabongo, D'Souza, and Auer}]{orkg-tdm}
Salomon Kabongo, Jennifer D'Souza, and S{\"o}ren Auer. 2021.
\newblock Automated mining of leaderboards for empirical ai research.
\newblock \emph{arXiv preprint arXiv:2109.13089}.

\bibitem[{Kakimpa et~al.(2012)Kakimpa, Hargreaves, and Owen}]{eng-kg}
B~Kakimpa, DM~Hargreaves, and JS~Owen. 2012.
\newblock An investigation of plate-type windborne debris flight using coupled
  cfd--rbd models. part ii: Free and constrained flight.
\newblock \emph{Journal of wind engineering and industrial aerodynamics},
  111:104--116.

\bibitem[{Kerrien et~al.(2007)Kerrien, Alam-Faruque, Aranda, Bancarz, Bridge,
  Derow, Dimmer, Feuermann, Friedrichsen, Huntley et~al.}]{intact}
Samuel Kerrien, Yasmin Alam-Faruque, Bruno Aranda, Iain Bancarz, Alan Bridge,
  Cathy Derow, Emily Dimmer, Marc Feuermann, Anja Friedrichsen, Rachael
  Huntley, et~al. 2007.
\newblock Intact—open source resource for molecular interaction data.
\newblock \emph{Nucleic acids research}, 35(suppl\_1):D561--D565.

\bibitem[{Kim et~al.(2003)Kim, Ohta, Tateisi, and Tsujii}]{genia}
J-D Kim, Tomoko Ohta, Yuka Tateisi, and Jun’ichi Tsujii. 2003.
\newblock Genia corpus—a semantically annotated corpus for bio-textmining.
\newblock \emph{Bioinformatics}, 19(suppl\_1):i180--i182.

\bibitem[{Kim et~al.(2011)Kim, Ohta, Pyysalo, Kano, and Tsujii}]{bionlp09}
Jin-Dong Kim, Tomoko Ohta, Sampo Pyysalo, Yoshinobu Kano, and Jun’ichi
  Tsujii. 2011.
\newblock Extracting bio-molecular events from literature—the bionlp’09
  shared task.
\newblock \emph{Computational Intelligence}, 27(4):513--540.

\bibitem[{Kim et~al.(2010)Kim, Medelyan, Kan, and Baldwin}]{kim2010semeval}
Su~Nam Kim, Olena Medelyan, Min-Yen Kan, and Timothy Baldwin. 2010.
\newblock Semeval-2010 task 5: Automatic keyphrase extraction from scientific
  articles.
\newblock In \emph{Proceedings of the 5th International Workshop on Semantic
  Evaluation}, pages 21--26.

\bibitem[{Kivim{\"a}ki et~al.(2012)Kivim{\"a}ki, Shipley, Allan, Sexton,
  Jokela, Virtanen, Tiemeier, Ebmeier, and Singh-Manoux}]{med-kg}
Mika Kivim{\"a}ki, Martin~J Shipley, Charlotte~L Allan, Claire~E Sexton, Markus
  Jokela, Marianna Virtanen, Henning Tiemeier, Klaus~P Ebmeier, and Archana
  Singh-Manoux. 2012.
\newblock Vascular risk status as a predictor of later-life depressive
  symptoms: a cohort study.
\newblock \emph{Biological psychiatry}, 72(4):324--330.

\bibitem[{Krallinger et~al.(2009)Krallinger, Izarzugaza, Rodriguez-Penagos, and
  Valencia}]{mutations}
Martin Krallinger, Jose~MG Izarzugaza, Carlos Rodriguez-Penagos, and Alfonso
  Valencia. 2009.
\newblock Extraction of human kinase mutations from literature, databases and
  genotyping studies.
\newblock \emph{BMC bioinformatics}, 10(8):1--20.

\bibitem[{Krallinger et~al.(2008)Krallinger, Leitner, Rodriguez-Penagos, and
  Valencia}]{biocreativeIIppi}
Martin Krallinger, Florian Leitner, Carlos Rodriguez-Penagos, and Alfonso
  Valencia. 2008.
\newblock Overview of the protein-protein interaction annotation extraction
  task of biocreative ii.
\newblock \emph{Genome biology}, 9(2):1--19.

\bibitem[{Krallinger et~al.(2010)Krallinger, Leitner, and
  Valencia}]{genedisease}
Martin Krallinger, Florian Leitner, and Alfonso Valencia. 2010.
\newblock Analysis of biological processes and diseases using text mining
  approaches.
\newblock \emph{Bioinformatics Methods in Clinical Research}, pages 341--382.

\bibitem[{Krallinger et~al.(2021)Krallinger, Miranda, Mehryary, Luoma, Pyysalo,
  and Valencia}]{drugprot}
Martin Krallinger, Antonio Miranda, Farrokh Mehryary, Jouni Luoma, Sampo
  Pyysalo, and Alfonso Valencia. 2021.
\newblock Drugprot shared task (biocreative vii track 1-2021) text mining
  drug-protein/gene interactions (drugprot) shared task.

\bibitem[{Krallinger et~al.(2017)Krallinger, Rabal, Akhondi, P{\'e}rez,
  Santamar{\'i}a, Rodr{\'i}guez, Tsatsaronis, Intxaurrondo, L{\'o}pez, Nandal,
  van Buel, Chandrasekhar, Rodenburg, L{\ae}greid, Doornenbal, Oyarz{\'a}bal,
  Lourenço, and Valencia}]{biocreativeVIIchemprot}
Martin Krallinger, Obdulia Rabal, Saber~Ahmad Akhondi, Mart{\'i}n~P{\'e}rez
  P{\'e}rez, Jesus Santamar{\'i}a, Gael~P{\'e}rez Rodr{\'i}guez, Georgios
  Tsatsaronis, Ander Intxaurrondo, Jos{\'e} Antonio~Baso L{\'o}pez, Umesh~K.
  Nandal, Erin~M. van Buel, Anjana Chandrasekhar, Marleen Rodenburg, Astrid
  L{\ae}greid, Marius~A. Doornenbal, Julen Oyarz{\'a}bal, An{\'a}lia Lourenço,
  and Alfonso Valencia. 2017.
\newblock Overview of the biocreative vi chemical-protein interaction track.

\bibitem[{Krallinger et~al.(2015)Krallinger, Rabal, Leitner, Vazquez, Salgado,
  Lu, Leaman, Lu, Ji, Lowe et~al.}]{chemdnerCorpus}
Martin Krallinger, Obdulia Rabal, Florian Leitner, Miguel Vazquez, David
  Salgado, Zhiyong Lu, Robert Leaman, Yanan Lu, Donghong Ji, Daniel~M Lowe,
  et~al. 2015.
\newblock The chemdner corpus of chemicals and drugs and its annotation
  principles.
\newblock \emph{Journal of cheminformatics}, 7(1):1--17.

\bibitem[{Krallinger et~al.(2011)Krallinger, Vazquez, Leitner, Salgado,
  Chatr-Aryamontri, Winter, Perfetto, Briganti, Licata, Iannuccelli
  et~al.}]{biocreativeiiippi}
Martin Krallinger, Miguel Vazquez, Florian Leitner, David Salgado, Andrew
  Chatr-Aryamontri, Andrew Winter, Livia Perfetto, Leonardo Briganti, Luana
  Licata, Marta Iannuccelli, et~al. 2011.
\newblock The protein-protein interaction tasks of biocreative iii:
  classification/ranking of articles and linking bio-ontology concepts to full
  text.
\newblock \emph{BMC bioinformatics}, 12(8):1--31.

\bibitem[{Krupp et~al.(2012)Krupp, Roussos, Kollmann, Paranicas, Mitchell,
  Krimigis, Rymer, Jones, Arridge, Armstrong et~al.}]{ast-kg}
N~Krupp, E~Roussos, P~Kollmann, C~Paranicas, DG~Mitchell, SM~Krimigis, A~Rymer,
  GH~Jones, CS~Arridge, TP~Armstrong, et~al. 2012.
\newblock The cassini enceladus encounters 2005--2010 in the view of energetic
  electron measurements.
\newblock \emph{Icarus}, 218(1):433--447.

\bibitem[{Li et~al.(2016)Li, Sun, Johnson, Sciaky, Wei, Leaman, Davis,
  Mattingly, Wiegers, and Lu}]{biocreativeVtaskcorpus}
Jiao Li, Yueping Sun, Robin~J Johnson, Daniela Sciaky, Chih-Hsuan Wei, Robert
  Leaman, Allan~Peter Davis, Carolyn~J Mattingly, Thomas~C Wiegers, and Zhiyong
  Lu. 2016.
\newblock Biocreative v cdr task corpus: a resource for chemical disease
  relation extraction.
\newblock \emph{Database}, 2016.

\bibitem[{Luan et~al.(2018)Luan, He, Ostendorf, and Hajishirzi}]{sciie}
Yi~Luan, Luheng He, Mari Ostendorf, and Hannaneh Hajishirzi. 2018.
\newblock Multi-task identification of entities, relations, and coreference for
  scientific knowledge graph construction.
\newblock In \emph{Proceedings of the 2018 Conference on Empirical Methods in
  Natural Language Processing}, pages 3219--3232.

\bibitem[{Ma and Hovy(2016)}]{Ma2016EndtoendSL}
Xuezhe Ma and Eduard~H. Hovy. 2016.
\newblock End-to-end sequence labeling via bi-directional lstm-cnns-crf.
\newblock \emph{CoRR}, abs/1603.01354.

\bibitem[{Martin et~al.(2012)Martin, Mooney, Dickinson, and West}]{agri-kg}
Sarah~L Martin, Sacha~J Mooney, Matthew~J Dickinson, and Helen~M West. 2012.
\newblock Soil structural responses to alterations in soil microbiota induced
  by the dilution method and mycorrhizal fungal inoculation.
\newblock \emph{Pedobiologia}, 55(5):271--281.

\bibitem[{Mendes et~al.(2011)Mendes, Jakob, Garc{\'\i}a-Silva, and
  Bizer}]{mendes2011dbpedia}
Pablo~N Mendes, Max Jakob, Andr{\'e}s Garc{\'\i}a-Silva, and Christian Bizer.
  2011.
\newblock Dbpedia spotlight: shedding light on the web of documents.
\newblock In \emph{Proceedings of the 7th international conference on semantic
  systems}, pages 1--8. ACM.

\bibitem[{Mitchell et~al.(2018)Mitchell, Cohen, Hruschka, Talukdar, Yang,
  Betteridge, Carlson, Dalvi, Gardner, Kisiel et~al.}]{mitchell2018never}
Tom Mitchell, William Cohen, Estevam Hruschka, Partha Talukdar, Bishan Yang,
  Justin Betteridge, Andrew Carlson, Bhanava Dalvi, Matt Gardner, Bryan Kisiel,
  et~al. 2018.
\newblock Never-ending learning.
\newblock \emph{Communications of the ACM}, 61(5):103--115.

\bibitem[{Mohan and Li(2018)}]{medmentions}
Sunil Mohan and Donghui Li. 2018.
\newblock Medmentions: A large biomedical corpus annotated with umls concepts.
\newblock In \emph{Automated Knowledge Base Construction (AKBC)}.

\bibitem[{Mondal et~al.(2021)Mondal, Hou, and Jochim}]{smallnlp}
Ishani Mondal, Yufang Hou, and Charles Jochim. 2021.
\newblock \href {https://doi.org/10.18653/v1/2021.findings-acl.165} {End-to-end
  construction of {NLP} knowledge graph}.
\newblock In \emph{Findings of the Association for Computational Linguistics:
  ACL-IJCNLP 2021}, pages 1885--1895, Online. Association for Computational
  Linguistics.

\bibitem[{Moro et~al.(2014)Moro, Cecconi, and Navigli}]{Moroetal:2014iswc}
Andrea Moro, Francesco Cecconi, and Roberto Navigli. 2014.
\newblock {M}ultilingual {W}ord {S}ense {D}isambiguation and {E}ntity {L}inking
  for {E}verybody.
\newblock In \emph{ISWC}, pages 25--28, Riva del Garda, Italy.

\bibitem[{Moro and Navigli(2015)}]{moro2015semeval}
Andrea Moro and Roberto Navigli. 2015.
\newblock Semeval-2015 task 13: Multilingual all-words sense disambiguation and
  entity linking.
\newblock In \emph{Proceedings of the 9th international workshop on semantic
  evaluation (SemEval 2015)}, pages 288--297.

\bibitem[{QasemiZadeh and Schumann(2016)}]{acl-rd-tec}
Behrang QasemiZadeh and Anne-Kathrin Schumann. 2016.
\newblock The acl rd-tec 2.0: A language resource for evaluating term
  extraction and entity recognition methods.
\newblock In \emph{Proceedings of the Tenth International Conference on
  Language Resources and Evaluation (LREC'16)}, pages 1862--1868.

\bibitem[{Schoch et~al.(2020)Schoch, Ciufo, Domrachev, Hotton, Kannan,
  Khovanskaya, Leipe, Mcveigh, O’Neill, Robbertse, Sharma, Soussov, Sullivan,
  Sun, Turner, and Karsch-Mizrachi}]{ncbi}
Conrad~L Schoch, Stacy Ciufo, Mikhail Domrachev, Carol~L Hotton, Sivakumar
  Kannan, Rogneda Khovanskaya, Detlef Leipe, Richard Mcveigh, Kathleen
  O’Neill, Barbara Robbertse, Shobha Sharma, Vladimir Soussov, John~P
  Sullivan, Lu~Sun, Seán Turner, and Ilene Karsch-Mizrachi. 2020.
\newblock \href {https://doi.org/10.1093/database/baaa062} {{NCBI Taxonomy: a
  comprehensive update on curation, resources and tools}}.
\newblock \emph{Database}, 2020.
\newblock Baaa062.

\bibitem[{Schubert(2006)}]{schubert2006turing}
Lenhart Schubert. 2006.
\newblock Turing's dream and the knowledge challenge.
\newblock In \emph{Proceedings of the national conference on artificial
  intelligence}, volume~21, page 1534. Menlo Park, CA; Cambridge, MA; London;
  AAAI Press; MIT Press; 1999.

\bibitem[{Shah et~al.(2003)Shah, Perez-Iratxeta, Bork, and
  Andrade}]{shah2003information}
Parantu~K Shah, Carolina Perez-Iratxeta, Peer Bork, and Miguel~A Andrade. 2003.
\newblock Information extraction from full text scientific articles: where are
  the keywords?
\newblock \emph{BMC bioinformatics}, 4(1):20.

\bibitem[{Soares et~al.(2014)Soares, Chandra, Thomas, Pedersen, Vallier, and
  Williams}]{bio-kg}
Filipa~AC Soares, Amit Chandra, Robert~J Thomas, Roger~A Pedersen, Ludovic
  Vallier, and David~J Williams. 2014.
\newblock Investigating the feasibility of scale up and automation of human
  induced pluripotent stem cells cultured in aggregates in feeder free
  conditions.
\newblock \emph{Journal of biotechnology}, 173:53--58.

\bibitem[{Tanabe et~al.(2005)Tanabe, Xie, Thom, Matten, and Wilbur}]{genetag}
Lorraine Tanabe, Natalie Xie, Lynne~H Thom, Wayne Matten, and W~John Wilbur.
  2005.
\newblock Genetag: a tagged corpus for gene/protein named entity recognition.
\newblock \emph{BMC bioinformatics}, 6(1):1--7.

\bibitem[{Unger et~al.(2014)Unger, Forascu, Lopez, Ngomo, Cabrio, Cimiano, and
  Walter}]{unger2014question}
Christina Unger, Corina Forascu, Vanessa Lopez, Axel-Cyrille~Ngonga Ngomo,
  Elena Cabrio, Philipp Cimiano, and Sebastian Walter. 2014.
\newblock Question answering over linked data (qald-4).

\bibitem[{Wang et~al.(2020)Wang, Shen, Huang, Wu, Dong, and
  Kanakia}]{wang2020microsoft}
Kuansan Wang, Zhihong Shen, Chiyuan Huang, Chieh-Han Wu, Yuxiao Dong, and
  Anshul Kanakia. 2020.
\newblock Microsoft academic graph: When experts are not enough.
\newblock \emph{Quantitative Science Studies}, 1(1):396--413.

\bibitem[{Wilkinson et~al.(2016)Wilkinson, Dumontier, Aalbersberg, Appleton,
  Axton, Baak, Blomberg, Boiten, da~Silva~Santos, Bourne
  et~al.}]{wilkinson2016fair}
Mark~D Wilkinson, Michel Dumontier, IJsbrand~Jan Aalbersberg, Gabrielle
  Appleton, Myles Axton, Arie Baak, Niklas Blomberg, Jan-Willem Boiten,
  Luiz~Bonino da~Silva~Santos, Philip~E Bourne, et~al. 2016.
\newblock The fair guiding principles for scientific data management and
  stewardship.
\newblock \emph{Scientific data}, 3(1):1--9.

\end{thebibliography}

\appendix

\section{Appendix}
\label{sec:appendix}

Supplementary to the Knowledge Graph in the Material Science domain presented in \autoref{sec:kg}, in Figures 13 to 20, we demonstrate nine examples of knowledge graphs for the respective nine remaining STEM domains we consider. In all graphs, nodes are color-coded by their concept type: orange corresponds to \textsc{process}, purple for \textsc{method}, green corresponds to \textsc{material}, and blue for \textsc{data}.


\begin{figure}[!htb]
\centering
\includegraphics[width=\linewidth]{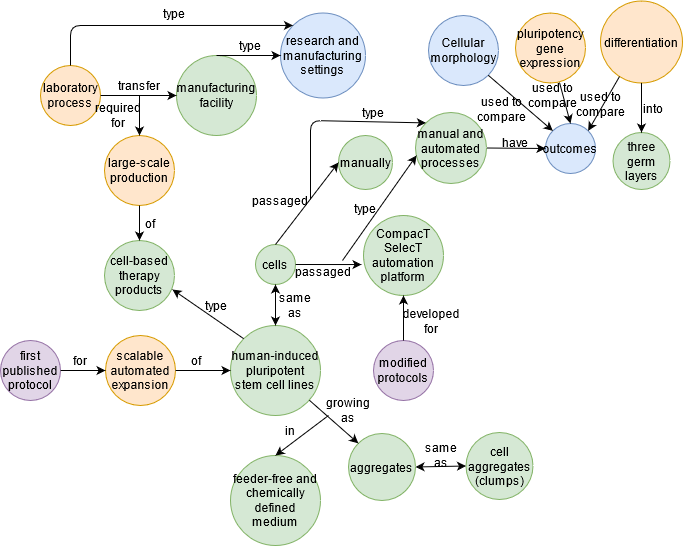}
\caption{Structured KG representation of a Biology domain publication Abstract~\cite{bio-kg} as \textsc{process}, \textsc{method}, \textsc{material}, and \textsc{data} entities.}
\end{figure}

\begin{figure}[!htb]
\centering
\includegraphics[width=0.8\linewidth]{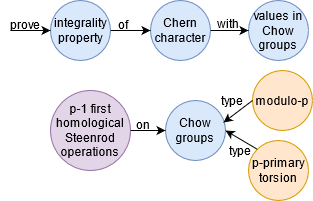}
\caption{Structured KG representation of a Mathematics domain publication Abstract~\cite{math-kg} as \textsc{process}, \textsc{method}, \textsc{material}, and \textsc{data} entities.}
\end{figure}

\begin{figure*}[!htb]
\centering
\includegraphics[width=\linewidth]{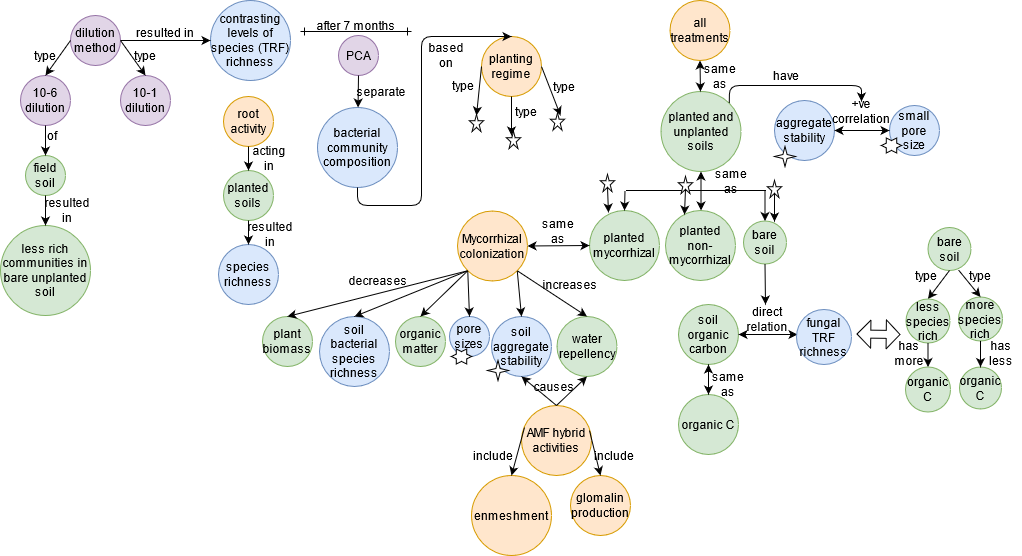}
\caption{Structured KG representation of an Agriculture domain publication Abstract~\cite{agri-kg} as \textsc{process}, \textsc{method}, \textsc{material}, and \textsc{data} typed entities.}
\end{figure*}

\begin{figure*}[!htb]
\centering
\includegraphics[width=\linewidth]{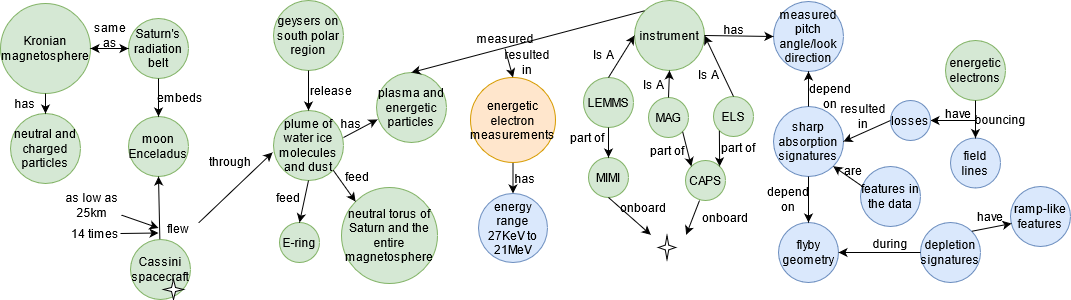}
\caption{Structured KG representation of an Astronomy domain publication Abstract~\cite{ast-kg} as \textsc{process}, \textsc{method}, \textsc{material}, and \textsc{data} typed entities.}
\end{figure*}

\begin{figure*}[!htb]
\centering
\includegraphics[width=\linewidth]{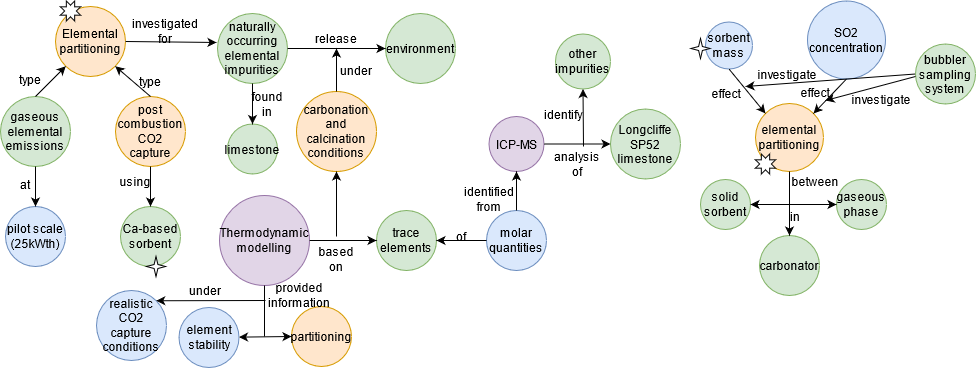}
\caption{Structured KG representation of a Chemistry domain publication Abstract~\cite{chem-kg} as \textsc{process}, \textsc{method}, \textsc{material}, and \textsc{data} typed entities.}
\end{figure*}

\begin{figure*}[!htb]
\centering
\includegraphics[width=\linewidth]{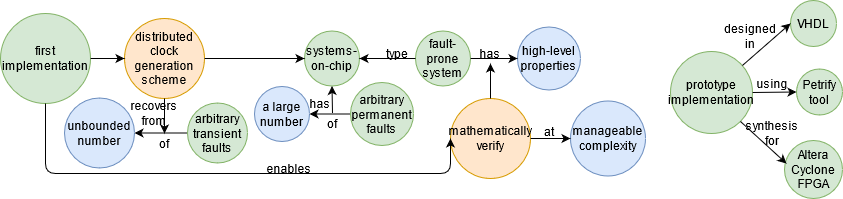}
\caption{Structured KG representation of a Computer Science domain publication Abstract~\cite{cs-kg} as \textsc{process}, \textsc{method}, \textsc{material}, and \textsc{data} typed entities.}
\end{figure*}

\begin{figure*}[!htb]
\centering
\includegraphics[width=\linewidth]{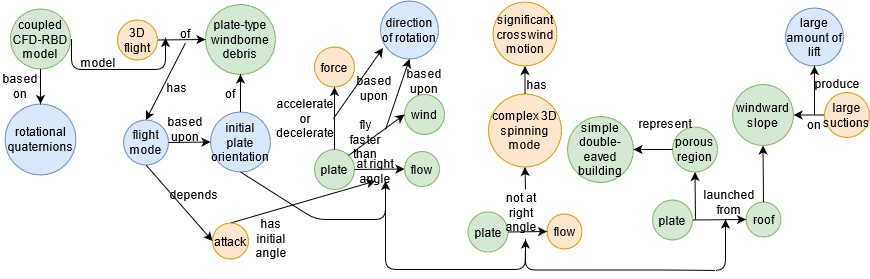}
\caption{Structured KG representation of an Engineering domain publication Abstract~\cite{eng-kg} as \textsc{process}, \textsc{method}, \textsc{material}, and \textsc{data} typed entities.}
\end{figure*}

\begin{figure*}[!htb]
\centering
\includegraphics[width=\linewidth]{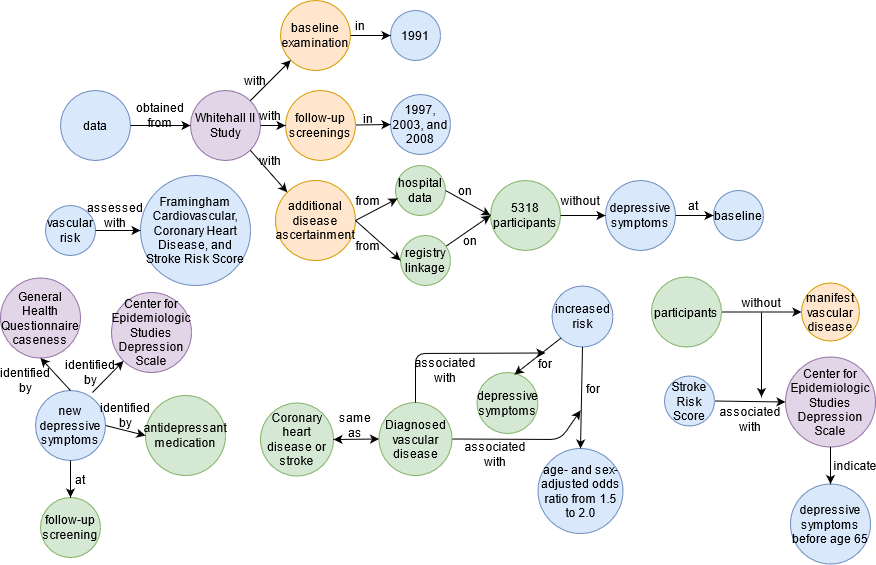}
\caption{Structured KG representation of a Medical domain publication Abstract~\cite{med-kg} as \textsc{process}, \textsc{method}, \textsc{material}, and \textsc{data} typed entities.}
\end{figure*}

\end{document}